\documentclass[runningheads]{llncs}

\usepackage{eccv}

\usepackage{soul}

\newcommand{\method}{CoFiMA\xspace}

\newcommand\etal{\emph{el~al.}}

\usepackage{eccvabbrv}

\usepackage{graphicx}
\usepackage{booktabs}
\usepackage{enumitem}
\usepackage{algorithm}
\usepackage{algorithmicx}
\usepackage{algpseudocode} 
\usepackage{amsmath}
\usepackage{maths}

\usepackage{arydshln} 
\definecolor{LightCyan}{rgb}{0.88,1,1}
\usepackage[dvipsnames]{xcolor}

\definecolor{keywordcolor}{rgb}{0.0, 0.0, 0.5} 
\definecolor{commentcolor}{rgb}{0.25, 0.5, 0.35} 
\definecolor{stringcolor}{rgb}{0.63, 0.13, 0.94} 

\usepackage{multirow}
\usepackage{bbding}
\usepackage{makecell}
\usepackage{colortbl} 

\usepackage[accsupp]{axessibility}  

\usepackage[pagebackref,breaklinks,colorlinks,citecolor=eccvblue]{hyperref}

\usepackage{orcidlink}

\begin{document}

\title{Weighted Ensemble Models Are Strong Continual Learners} 

\titlerunning{Weighted Ensemble Models Are Strong Continual Learners}

\author{Imad Eddine Marouf$^{1}$\orcidlink{0000-0003-2479-0564}
\and
Subhankar Roy$^{2}$\orcidlink{0009-0008-2395-8111}
\and
Enzo Tartaglione$^{1}$\orcidlink{0000-0003-4274-8298}
\and \\
Stéphane Lathuilière$^{1}$\orcidlink{0000-0001-6927-8930}
}

\institute{$^{1}$Télécom-Paris, Institut Polytechnique de Paris, $^{2}$University of Trento, Italy\\
\email{$^{1}$\{first.lastname\}@telecom-paris.fr}\\
\email{$^{2}$\{first.lastname\}@unitn.it}
}
\authorrunning{I. Marouf et al.}

\maketitle

\begin{abstract}
  In this work, we study the problem of continual learning (CL) where the goal is to learn a model on a sequence of tasks, under the assumption that the data from the previous tasks becomes unavailable while learning on the current task data. CL is essentially a balancing act between learning on the new task (\ie plasticity) and maintaining the performance on the previously learned concepts (\ie stability). To address the stability-plasticity trade-off, we propose to perform weight-ensembling of the model parameters of the previous and current tasks. This weighted-ensembled model, which we call \underline{\textbf{Co}}ntinual \underline{\textbf{M}}odel \underline{\textbf{A}}veraging (or \textbf{CoMA}), attains high accuracy on the current task by leveraging plasticity, while not deviating too far from the previous weight configuration, ensuring stability. We also propose an improved variant of CoMA, named \underline{\textbf{Co}}ntinual \underline{\textbf{Fi}}sher-weighted \underline{\textbf{M}}odel \underline{\textbf{A}}veraging (or \textbf{CoFiMA}), that selectively weighs each parameter in the weights ensemble by leveraging the Fisher information of the weights of the model. Both variants are conceptually simple, easy to implement, and effective in attaining state-of-the-art performance on several standard CL benchmarks. Code is available at: \href{https://github.com/IemProg/CoFiMA}{https://github.com/IemProg/CoFiMA}.
\end{abstract}

\section{Introduction}
\label{sec:intro}
Continually learning from a sequence of tasks with a unified model is a challenging problem due to \textit{catastrophic forgetting} (CF)~\cite{french1999catastrophic} -- a phenomenon that is marked by deterioration of performance on previously seen data. Continual learning (CL) has emerged as a solution to CF that allows models to assimilate information from new tasks while retaining classification capability for the previously learned classes~\cite{masana2022class}.
Until recently, CL approaches predominantly focused on relatively small networks, often ResNets~\cite{he2015residual}, starting from random initialization~\cite{masana2022class,wang2023comprehensive}. 
Lately, the prominence of large Pre-Trained Models (PTMs)~\cite{d2021convit, dosovitskiy2021image, radford2021learning} -- Vision Transformer (ViT)~\cite{dosovitskiy2021image, liu2021swin} pre-trained on large datasets (\eg ImageNet~\cite{ridnik2021imagenet21k}, LAION-400M~\cite{schuhmann2021laion}) -- has led to an influx of CL methods that are leveraging the strong representation of the PTMs, causing a paradigm shift in CL~\cite{wang2022dualprompt, wang2022learning, wang2022s,villa2022pivot, AdamAdapter}.
In detail, numerous PTM-based CL methods~\cite{wu2022class,effect_scale_ptms,mehta2023empirical} have empirically validated that a good initial representation, obtained with the help of large and diverse pre-training, facilitates incremental learning since new tasks can be learned with fewer training steps. However, sequential full fine-tuning of the PTM backbone results in deterioration of the original PTM representation, alongside significant forgetting on the previously learned tasks~\cite{SLCA, mcdonnell2024ranpac, AdamAdapter,panos2023first}. To counteract overfitting, many methods have been proposed that either heuristically confine the PTM fine-tuning only to the first adaptation session~\cite{mcdonnell2024ranpac,AdamAdapter,panos2023first} or carefully choose a low learning rate to fine-tune the backbone~\cite{SLCA}. Nevertheless, achieving a \textit{satisfactory balance between quickly accruing knowledge on new tasks while preserving the generalizability of the PTMs} remains an open research question.

\begin{figure}[t]
    \begin{center}
    \includegraphics[width=\columnwidth]{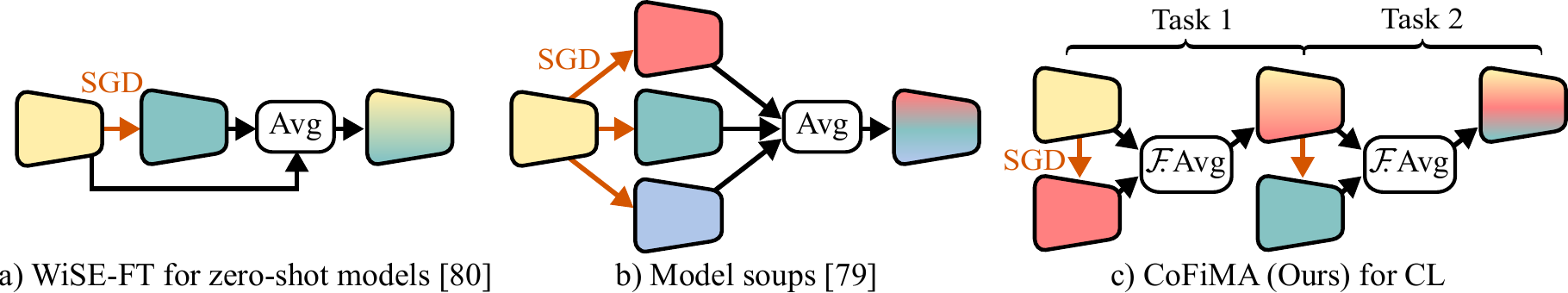}
    \end{center}
    \caption{\textbf{Comparison of existing model averaging techniques with our proposed technique for CL}. (a) Averaging the weights of the pre-trained model and the one fine-tuned leads to simultaneous improvement in out-of-distribution and target dataset performance. (b) Models soups combines multiple fine-tuned models resulting in a robust unified model. (c) In the proposed CoFiMA the weights of the current and past models are weighted based on their Fisher Information matrices (represented by $\mathcal{F}$), resulting in balanced performance for both the current and old tasks.
    }
    \label{fig:teaser}
\end{figure}

In the quest for achieving robust fine-tuning of PTMs, several studies have investigated the application of weight averaging (WA) methods~\cite{murata2022learning, Kim_2023_CVPR, abraham2005memory,wortsman2021robust,wortsman2022model,matena2022merging}. The essence of these methods is to ensemble several fine-tuned PTMs to obtain a single model that encapsulates the representational capabilities of multiple models. While, to our knowledge, WA has not yet been investigated for CL, two recent studies~\cite{wortsman2021robust,wortsman2022model} have sparked our interest in examining the feasibility of WA for CL.
Firstly, WiSE-FT~\cite{wortsman2021robust} improves the robustness of the fine-tuning procedure of zero-shot classifiers like CLIP by averaging the weights of both fine-tuned and pre-trained models (Fig.~\ref{fig:teaser}(a)). Here, the resulting ensemble exhibits high accuracy on the target distribution while preserving the out-of-distribution (OOD) performance of the original PTM. It is important to recognize that model robustness and CL intersect: the balance between \textit{OOD performance and target performance} mirrors the CL \textit{stability-plasticity trade-off}, where the goal is to accommodate new tasks while preserving efficacy on prior tasks~\cite{murata2022learning, Kim_2023_CVPR, abraham2005memory}. Secondly, ``Model soups''~\cite{wortsman2022model} shows that combining multiple fine-tunings of the same PTM through WA enhances performance on both in- and out-of-distribution tasks (Fig.~\ref{fig:teaser}(b)). Their experiments underscore the potential of WA with a relatively large pool of models (\eg 32), mirroring the number of tasks typically encountered in CL. Nevertheless, Model soups do not establish the feasibility of using WA when models are trained on different tasks, a capability required for CL.

Motivated by these observations, \textit{we cast PTM-based CL as a robust fine-tuning problem} and propose \underline{\textbf{Co}}ntinual \underline{\textbf{M}}odel \underline{\textbf{A}}veraging (\textbf{CoMA}) as a response to the PTM fine-tuning conundrum in CL. In CoMA we linearly combine the weights from the previous task $~{t\!-\!1}$ with the weights of the fully fine-tuned model from the current task $t$ (Fig.~\ref{fig:teaser}(c)). Full fine-tuning on the current task ensures plasticity, while ensembling the weights with the previous task checkpoint preserves performance on the previous task. The ensemble at task $t$ is then used as an initialization for the next task $~{t\!+\!1}$. In essence, our proposed CoMA exploits the model averaging techniques~\cite{wortsman2021robust,wortsman2022model} to the CL scenario by extending the model averaging to a sequence of downstream tasks. Intuitively, CoMA is effective because task-specific models optimized sequentially could lie in the same basin of the total loss landscape~\cite{mirzadeh2020linear, neyshabur2021transferred}. 

The model averaging in CoMA assumes that all the network's weights have the same importance for a given task. While effective, putting equal importance to all weights could result in a weight-ensembled model that lies in an error basin with high loss~\cite{wortsman2022model}. To mitigate this problem, we aim to selectively ensemble the weights based on the importance of each weight parameter for a given task. Inspired by Elastic Weight Consolidation (EWC)~\cite{kirkpatrick2017overcoming}, and the more recent work of Matena \etal~\cite{matena2022merging}, we leverage the Fisher information~\cite{bayesrules, spall2005monte} to weigh the model parameters during model averaging. Fisher information inherently captures the importance of each weight parameter on the dataset (or task) the model has been trained on. We call this variant of CoMA as \underline{\textbf{Co}}ntinual \underline{\textbf{Fi}}sher-weighted \underline{\textbf{M}}odel \underline{\textbf{A}}veraging (\textbf{CoFiMA}), and shown in Fig.~\ref{fig:teaser}(c). CoFiMA stores only the Fisher values of the previous task, thereby rendering Fisher-weighted averaging compatible with CL constraints. \method maintains computational efficiency, necessitating only a singular forward and backward pass on the data of the current task to estimate Fisher values~\cite{pascanu2014revisiting, kirkpatrick2017overcoming} and yields state-of-the-art performance surpassing both CoMA and existing PTM-based CL solutions.

Our \textbf{contributions} are summarised as follows:
\begin{itemize}
    \item We draw a parallel between \textit{robust fine-tuning} and \textit{continual learning} and show that model-averaging is a simple yet effective solution for \textit{PTM-based} CL problem. We propose CoMA, a weight-ensemble inspired CL approach, that addresses the challenging task of stability-plasticity trade-off.
    \item We extend CoMA to \method by employing Fisher information to adaptively weigh the parameters of the previous and current task model.
    \item We run extensive experiments on several standard CL benchmarks and demonstrate that \method yet being simple, it consistently outperforms PTM-based CL solutions.
\end{itemize}

 \section{Related Work}

\textbf{Continual Learning with PTM. }
Not a long time ago, the predominant focus in CL has been on the sequential training of deep neural networks from scratch, aiming to proficiently acquire new tasks while mitigating forgetting of preceding tasks. Typical CL strategies encompass \textit{regularization}-based approaches~\cite{kirkpatrick2017overcoming,aljundi2018memory,zenke2017continual,li2017learning,dhar2019learning}, which maintain the initial model and selectively stabilize parameter or prediction alterations; \textit{replay}-based approaches~\cite{wang2021ordisco,wu2019large,prabhu2020gdumb,buzzega2020dark}, that seek to approximate and regenerate previously learned data distributions; and \textit{architecture}-based approaches~\cite{yang2022continual, serra2018overcoming,rusu2016progressive}, which allocate discrete parameter sub-spaces for each task. 

Differently, the recent trajectory of CL research has probed into the advantages of PTMs~\cite{wang2022learning, wang2022dualprompt, effect_scale_ptms}. Representations derived from pre-training have demonstrated the capacity to facilitate not only knowledge transfer but also resilience against catastrophic forgetting during downstream continual learning~\cite{ramasesh2021effect,mehta2023empirical}. Moreover, learning on substantial base classes during the pre-training phase permits CL with minimal adaptations~\cite{wu2022class}. For example, L2P~\cite{wang2022learning} leveraged techniques inspired by pre-trained knowledge utilization in NLP, employing an additional set of learnable parameters, termed ``prompts'', which guide a pre-trained representation layer in learning incremental tasks. DualPrompt~\cite{wang2022dualprompt} elaborated on this concept by attaching supplementary prompts to the pre-trained representation layer to facilitate the learning of both task-invariant and task-specific instructions. Though prompt-based approaches have been documented to significantly outperform conventional CL baselines, they introduce an extra inference cost. Recently, Zhang~\etal~\cite{SLCA} showed that sequential fine-tuning with a small learning rate using PTMs outperforms traditional CL approaches. Wang~\etal~\cite{wang2022coscl, Wang_2023} propose an architecture that employs multiple narrower sub-networks to manage incremental tasks,  effectively reducing generalization errors in CL. However, this approach introduces increased complexity.

Unlike prior approaches in CL that enhance PTMs using prompts~\cite{wang2022dualprompt, wang2022learning}, ensemble of experts~\cite{wang2022coscl, Wang_2023, divideforgetensembleselectively} or replay-buffers~\cite{wang2021ordisco,wu2019large,prabhu2020gdumb,buzzega2020dark}, \method employs a different strategy. CoFiMA enables plasticity by unlocking all model parameters to be fine-tuned unlike ~\cite{mcdonnell2024ranpac, wang2022learning, wang2022dualprompt}. Furthermore, it reduces forgetting during training by averaging the weights of parameters from previous models. In contrast to EWC~\cite{kirkpatrick2017overcoming}, which uses the Fisher Information Matrix (FIM) as a regularization term for L2-transfer between tasks, \method applies the FIM as a weighting factor to assess the significance of weights for each task, without any regularization constraints (more details in supplementary material).  

\noindent\textbf{Output-space/weight-space ensembles. }Traditional ensemble methods, or output-space ensembles, combine multiple classifiers' predictions, often outperforming single models and providing more calibrated uncertainty estimates under distribution shifts~\cite{friedman2001elements, deepensembles, ovadia2019can, stickland2020diverse,matena2022merging}. These output-space ensembles, however, demand substantial computational resources at inference. Weight-space ensembles offer a computationally efficient alternative by interpolating between model weights~\cite{wortsman2021robust, wortsman2022model, izmailov2018averaging, nichol2018first, szegedy2016rethinking}. Wortsman~\etal~\cite{wortsman2021robust} achieved this by interpolating between zero-shot CLIP and fine-tuned model weights, resulting in performance gains on both the fine-tuning task and under-distribution shifts. Matena~\etal~\cite{matena2022merging} propose an advanced WA technique that uses Fisher-values for different text classification tasks but they do not investigate the feasibility of this approach for CL. 
In Federated Learning (FL), model averaging, notably FedAvg, is a fundamental technique for amalgamating insights from decentralized data while upholding privacy~\cite{mcmahan2023communicationefficient}. This approach involves training local models on distributed nodes and averaging their parameters to update a global model, thus enhancing learning efficiency and data privacy~\cite{mcmahan2023communicationefficient, 9084352, kairouz2021advances}. Our approach differs from existing WA approaches~\cite{wortsman2021robust, wortsman2022model, matena2022merging} as it focuses on sequential finetuning with weight-averaging to be accustomed to the CL setting, we iteratively perform our procedure once per task, using the averaged model at each task as the initialization for the next task.

WA is also strongly related to the notion of linear model connectivity, introduced by Frankle~\etal~\cite{frankle2020linear}. This concept identifies a condition where the accuracy levels are maintained throughout the linear interpolation between the weights of two separate networks. Interpolation of neural network weights has been shown to maintain high accuracy across various scenarios, along a shared optimization trajectory~\cite{frankle2020linear,wortsman2021robust,entezari2021role,choshen2022fusing,izmailov2018averaging,wortsman2022model}.  Analogously, Neyshabur~\etal~\cite{neyshabur2021transferred} demonstrates that there exists a connection between minima obtained by pre-trained models versus freshly initialized ones. They note that there is no performance barrier between solutions coming from pre-trained models, but there can be a barrier between solutions of different randomly initialized models. Mizradeh~\etal~\cite{mirzadeh2020linear} investigated linear connectivity in the context of multi-task learning and CL. They show that there exists a linear path solution between two models trained on two tasks ``A'' and ``B'': one excelling in task A, and the other fine-tuned on both tasks A and B.  These works offer a solid foundation of both theoretical and empirical evidence supporting the efficacy of linear interpolation in model performance. Building on these findings, our work presents a novel solution tailored to CL, an area yet to be explored.

 \section{Method}
\label{method}

\subsection{Problem formulation and overview}
\textbf{Continual learning with pre-trained models. } We consider a classification model \( M_{\varphi}(\cdot)\!=\!h(f(\cdot)) \), where \( f(\cdot) \) is a feature extractor and \( h(\cdot) \) is a classification head, both parameterized by a unified set of parameters \( \varphi \). The feature extractor \( f_{\thetavect} \) is initialized with parameters $\thetavect_0$ of a PTM and then trained sequentially on a series of incremental tasks, each represented by the corresponding training set $\mathcal{D}_t$, for \( t \in \{1, \ldots, T\} \). The primary objective is to achieve robust performance across the test sets associated with these tasks. Specifically, for each task \( t \), the dataset \( \mathcal{D}_{t} \) is defined as 
$ \mathcal{D}_t = \bigcup_{c \in C_t} \mathcal{D}_t^c $
where $\mathcal{D}_{t}^c = { (\xmat_{n}^{c},\yvect_{n}^{c})}_{n=1}^{N_c}$, and $C_t$ denotes the set of novel classes introduced in task $t$. Here, $N_c$ represents the number of training instances for each class $c$, with \( (\xmat_n^c, \yvect_n^c) \) denoting the $n$-th training instance and its corresponding label. In Class-Incremental Learning (CIL), evaluation is conducted across all the observed classes without the need for task-index labels~\cite{vandeven2019scenarios}. 


This problem poses two primary challenges: (i) the necessity to adapt the knowledge acquired from the PTM to new tasks; and (ii) the importance of maintaining the model's comprehensive learning capabilities to avoid forgetting previously acquired knowledge while assimilating new tasks.

\noindent \textbf{Overview.} In this work we propose \textit{model averaging} as an effective solution for PTM-based CIL. As fine-tuning a PTM on a new task causes the weights to deviate from the original PTM and previous task configuration, model averaging avoids forgetting on a previous task and maintains the generalizability of PTM by averaging the weights of the models of the previous task and current task. Each task concludes with inference on all tasks seen so far using the averaged model and using it as an initialization for the next one. We name this approach \textit{Continual Model Averaging} (CoMA) and will be discussed in detail in Sec.~\ref{sec:coma}.


We extend CoMA by introducing Continual Fisher-weighted Model Averaging (CoFiMA), wherein we average two models at any task $t$ based on an additional weighing coefficient that is determined by Fisher information~\cite{spall2005monte, bayesrules}. Fisher information of a given model parameter dictates the importance of that particular weight towards a task. This procedure is detailed in Sec.~\ref{sec:cofima}.

    \begin{figure*}[t]
     \centering
     \includegraphics[width=\linewidth]{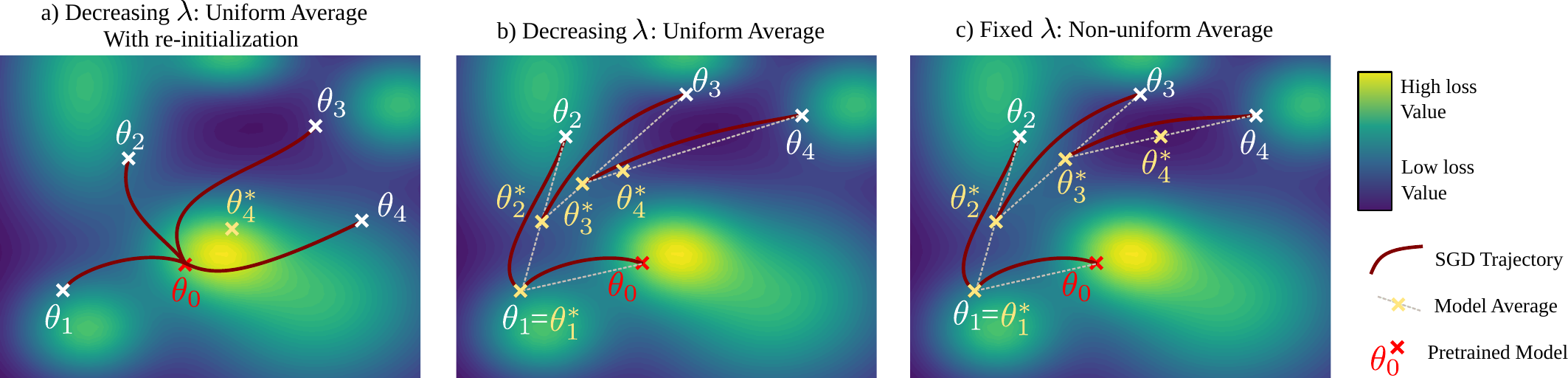}
             \caption{Illustration of the parameter trajectory with model averaging in the loss landscape. \textbf{(a)} The trajectory where models are re-initialized at the start of each task leading to disparate solutions. \textbf{(b)} Depicts decreasing $\lambda\!=\!1/t$: Uniform Averaging without re-initialization, showing the convergence of model parameters towards a solution that balances between tasks. \textbf{(c)} Recent tasks are given more weight, resulting in a solution that remains close to the latest task's model while considering previous tasks.}
       \label{fig:pipeline}
     \end{figure*} 

\subsection{Continual Model Averaging (CoMA)}
\label{sec:coma}

Temporarily disregarding the typical memory limitations of CIL, a practical approach through model ensembling is proposed as follows: individual models are trained for each task sequentially, starting from the pre-trained model, and the model is saved after each task. Suppose we have trained on a sequence of \( T \) tasks: this method will result in \( T \) distinct networks, each with its own set of parameters, denoted as \( \thetavect_1, \ldots, \thetavect_T \). The objective is to create a composite neural network with parameters \( \thetavect \) ensuring good performance across all the tasks.

We model the posterior over the composite parameters $\thetavect$, conditioned on the task-specific parameters $\thetavect_t$, denoted by \( p(\thetavect|\thetavect_t) \), as an isotropic Gaussian distribution \( \thetavect \sim \mathcal{N}(\thetavect_t, \Imat) \), where $\Imat$ is the identity matrix~\cite{wortsman2021robust,matena2022merging}.
 We treat the models $\thetavect_1, \ldots, \thetavect_T$ as independent observations of the composite model $\thetavect$ and maximizing the log-likelihoods of the posterior distribution of $\thetavect$ over all \( T \) the tasks leads to the following optimization problem:
\begin{equation}
\thetavect^* = \arg\max_{\thetavect} \frac{1}{T} \sum_{t=1}^{T} \log p(\thetavect|\thetavect_t).
\end{equation}
The solution to this optimization problem has closed-form and is the simple average of the model parameters~\cite{mcmahan2023communicationefficient, matena2022merging}. With a slight abuse in notation, where we use the sum operator to denote the element-wise summation across sets, it can be written:
\begin{equation}
\label{eq:avg}
\thetavect^* = \frac{1}{T} \sum_{t=1}^{T} \thetavect_t .
\end{equation}

In addition to the support of the well-established likelihood maximization framework, averaging models as in Eq.~\eqref{eq:avg} gains further validation from the insights provided by Mirzadeh~\etal~\cite{mirzadeh2020linear}. Their research shows that when two models are each trained on distinct tasks, a model adept in both tasks often exists within the linear interpolation of their parameter spaces.
Below, we detail how we adapt this model-averaging method to the constraints of CIL where the task data arrives sequentially, and without storing previous data. 

First, initiating the training of each task from the same pre-trained model $\thetavect_{0}$ can result in convergence to distinct regions within the parameter space~\cite{frankle2020linear, ramé2023diverse}. This case is illustrated in Fig.~\ref{fig:pipeline}(a). If the learned parameters are too distant, the Gaussian assumption of the posterior distribution is no longer valid, and averaging the network could lead to high-loss regions and poor performance of the aggregated model (see Sec.~\ref{sec.ablation}). 
Therefore, for each task $t$, we initiate finetuning from \(\thetavect^*_{t-1}\) rather than the initial pre-trained model $\thetavect_0$. This change limits the risk of reaching distant parameter regions. 

Second, one of the goals of our approach is to prevent the number of models from growing linearly with $t$. Therefore, we shift from simultaneous averaging all models \( \thetavect_1, \ldots, \thetavect_{T}\) to the iterative computation of the average \( \thetavect^*_t \) at each task $t$:
\begin{equation}
\label{eq:update}
\thetavect^{*}_{t} = \lambda \thetavect_{t} + (1 - \lambda)\thetavect^*_{t-1},
\end{equation}
where $\lambda\!\in\![0,1]$. To initialize this recursion, we start at $t\!=\!1$ with $\thetavect^*_{1}$ which is set to the model parameters obtained at the end of the first task. Note that, when $\lambda=\frac{1}{t}$, Eq.~\eqref{eq:update} is strictly equivalent to the average in Eq.~\eqref{eq:avg} (see supplementary material), aligning with the maximum-likelihood solution and adhering to our initial hypothesis of isotropic Gaussian posteriors. However, our experiments highlight that such a parameter choice might result in suboptimal performances and that assigning variable weights to each model proves advantageous. This phenomenon might be attributed to the possibility of encountering a region with a high loss value in between all the task-specific models (see Fig.~\ref{fig:pipeline}(b)).
Therefore, we propose to perform a non-uniform averaging giving higher importance to the latest tasks as shown in Fig.~\ref{fig:pipeline}(c). This is obtained by using a constant weight parameter $\lambda$. The motivation for giving more importance to the latest tasks is that early models are trained only on the first few tasks, while the last model has encoded the knowledge from both old and recent tasks via sequential fine-tuning (see supplementary material). In this way, it is expected that a better trade-off would exist along this trajectory, and the two endpoints $\thetavect_{t-1}$ and $\thetavect_{t}$ are smoothly connected without significant loss barrier or performance drop along the path~\cite{frankle2020linear}.

In terms of memory, during each training phase, the storage overhead of our approach compared to naive sequential fine-tuning over all the tasks is limited to the size of a single model. However, when transitioning to subsequent tasks, only $\thetavect^{*}_{t}$ requires storage.

\noindent\textbf{Handling Classifier Parameters. }
In each novel task $t$, there are unique parameters (specifically, new head parameters associated with new classes) not present in the preceding models that are subject to averaging. To accommodate this, we restrict the averaging process (as in Eq.~\eqref{eq:update}) exclusively to the parameters that are common across models (this includes both backbone parameters and the shared portions of head parameters) while excluding the new head weights (pertaining to new classes) from the averaging.

\subsection{Continual Fisher-weighted Model Averaging (\method)}
\label{sec:cofima}
Uniform weight-averaging operates under the implicit assumption that all parameters of the model have the same importance for the training task $t$. 
This assumption could potentially compromise model performance since different parameters can have various impacts on the network output~\cite{kirkpatrick2017overcoming,matena2022merging}. To enhance the averaging process, we propose a more refined model averaging for CIL. Specifically, we employ 
the Fisher information matrix~\cite{spall2005monte, bayesrules}, which encapsulates the quantity of information that observed data $\xmat$ provides about the network parameters $\thetavect_t$. The Fisher information matrix $F_{\thetavect_t}$ of a neural network $p_{\thetavect_t}(\yvect|\xmat)$ trained to predict an output $\yvect$ from input data $\xmat$ is computed as in~\cite{NIPS1996_39e4973b}:
\begin{align}
\Fmat_{\thetavect_t} = \mathbb{E}_{\xmat {\sim} \mathcal{D}_t} \big[ \mathbb{E}_{\yvect \sim p(\yvect|\xmat, \thetavect_t)} [ &\nabla_{\thetavect_t} \log p(\yvect|\xmat,\thetavect_t)\nabla_{\thetavect_t} \log p(\yvect|\xmat,\thetavect_t)^{T} ] \big].
\end{align}
Given the computational cost of storing the full Fisher matrix, this study adopts the \textit{diagonal} of the Fisher information matrix for practicality~\cite{pascanu2014revisiting, spall2005monte, kirkpatrick2017overcoming}. The computation of the diagonal Fisher is feasible within the same order of complexity as the standard back-propagation training process, as it necessitates only a single gradient computation per data point, corresponding to an extra epoch at the end of each task training. Through Monte-Carlo sampling over $\xmat$~\cite{spall2005monte, 4586850}, the diagonal Fisher matrix can be estimated as follows:
\begin{equation}
\hat{F}_{\thetavect_t} = \frac{1}{N} \sum_{i=1}^{N} \mathbb{E}_{\yvect \sim p(\yvect|\xmat_i,\thetavect_t)} \left[ [\nabla_{\thetavect_t} \log p(\yvect|\xmat_i,\thetavect_t)]^2 \right],
\end{equation} 
where the expectation over $\yvect$ is estimated from the data samples $N$.
The estimation of this Fisher information matrix \(F_{\thetavect}\) is based on the assumption that the statistical properties of the data are locally similar around the parameter values~\cite{soen2021variance, 8493265}. This assumption aligns with the Gaussian assumption previously utilized in deriving CoMA. Furthermore, as discussed in the work of Chizat~\etal~\cite{chizat2020lazy}, the use of PTMs limits the risk of facing significant changes in parameter distributions which would violate the \textit{locality assumption}~\cite{bayesrules}.

We consider that the posterior of each task model \( p(\thetavect|\thetavect_t) \) is defined as $~{\thetavect \!\sim\!\mathcal{N}({\thetavect}_t, \Fmat_{t}^{-1})}$. As in Eq.~\eqref{eq:avg}, maximization of the average of the task-specific log-likelihoods brings us to the following closed-form solution:
\begin{equation}
\thetavect^{*} = \frac{\sum_{t=1}^{T}  \Fmat_{t} \thetavect_{t}}{\sum_{t=1}^{T} \Fmat_{t}} ,
\end{equation}
with a slight abuse in notation, where we use the division operator to denote the element-wise division.
As in Sec.~\ref{sec:coma}, we adapt this formulation making it iterative and increasing the importance of the latest task in the parameter update. This brings us to the following update rule for $\theta_t^{*}$:
\begin{equation}
\thetavect_{t}^{*} = \frac{\lambda \Fmat_{t} \thetavect_{t} + (1- \lambda) \Fmat_{t-1} \thetavect^{*}_{t-1}}{\lambda \Fmat_{t} +(1 - \lambda) \Fmat_{t-1}},
\label{eq:main}
\end{equation}
where $F_{t-1}$ is the Fisher matrix estimated with $\thetavect^{*}_{t-1}$ on the data $\mathcal{D}_{t-1}$. In terms of memory requirements, this solution requires the storage of only the model parameters $\thetavect^{*}_{t}$ and the Fisher matrix $\Fmat_{t-1}$ in between two consecutive tasks $t$ and $t\!-\!1$. Similarly to CoMA, the recursion is initialized at the end of the first task with $\thetavect^*_1\!=\!\thetavect_1$ and estimating $\Fmat_1$ with $\thetavect_1$ on $\mathcal{D}_1$.
 \section{Experiments}
\label{sec:exp}
In this section, we first briefly describe the experimental setups and then present the experimental results. 
\subsection{Experimental Setups}
\textbf{Datasets and Settings.} We use PILOT~\cite{sun2023pilot} framework for our experiments. We conduct experiments on four CIL benchmarks: CIFAR-100~\cite{krizhevsky2009learning}, ImageNet-R~\cite{hendrycks2021many}, CUB-200~\cite{wah2011caltech}, and Cars-196~\cite{krause20133d}. The CIFAR-100 dataset~\cite{krizhevsky2009learning} has 100 classes of natural images, each with 500 training images. The ImageNet-R dataset~\cite{hendrycks2021many} includes images from 200 classes, divided into 24,000 for training and 6,000 for testing. These images, although related to ImageNet-21K, are considered challenging for the PTM because they are either hard examples from ImageNet or new images in different styles. The CUB-200 dataset~\cite{wah2011caltech} consists of images from 200 bird classes, with about 60 images per class, half for training and half for testing. The Cars-196 dataset~\cite{krause20133d} is made up of 196 types of car images, split into 8,144 for training and 8,040 for testing, maintaining a similar class ratio.  
The first two focus on fine-grained classifications, while the last two datasets (\ie, CIFAR-100, and ImageNet-R) are standard benchmarks for CL. Following SLCA~\cite{SLCA} we split each benchmark into 10 tasks. We report the results in the class-incremental setting, \ie the task id is not known during inference.

\noindent \textbf{Metrics.} We report the average classification accuracy of all the classes ever seen after learning each incremental task (denoted as \textit{Inc-Acc} (\%)) and the accuracy after learning the last task (denoted as \textit{Last-Acc} (\%)).

\noindent\textbf{Baselines and Competitors.} We compare with the state-of-the-art PTM-based CIL methods L2P~\cite{wang2022learning}, DualPrompt~\cite{wang2022dualprompt}, SLCA~\cite{SLCA}, and RanPAC~\cite{mcdonnell2024ranpac}. We also utilize the same PTM as the initialization for classical CL methods GDumb~\cite{prabhu2020gdumb}, LwF~\cite{li2017learning}, DER~\cite{buzzega2020dark},
BIC~\cite{wu2019large}, and EWC~\cite{kirkpatrick2017overcoming}. Additionally, we
report the following baselines: sequentially fine-tuning of the model (denoted as Seq FT), and Prototype-classifier~\cite{janson2023simple}, which is a cosine similarity classifier on the extracted features of the PTM. Joint-Training is an upper bound, where the model has been trained on all the tasks at the same time.

\noindent\textbf{Implementation details.} We adopted two kinds of PTMs in our experiments: a ViT-B/16 \cite{dosovitskiy2021image} backbone supervisedly pre-trained on ImageNet-21K \cite{ridnik2021imagenet21k}, the default PTM unless otherwise stated; and ViT-B/16 backbone with self-supervised pre-training using MoCo-V3 \cite{chen2021mocov3} on ImageNet-1K.  We follow the implementation of SLCA~\cite{SLCA} that adapts a small learning rate of 0.0001 for the representation layer and 0.01 for the classification layer, along with the class-alignment strategy. 
We set the batch size to 128, \textit{$\lambda\!=\!0.4$} for supervised and \textit{$\lambda\!=\!0.2$} for self-supervised pre-training, in all our experiments.

\begin{table}[t]
    \centering
    \caption{State-of-the-art comparison on CUB-200, Cars-196, CIFAR-100, and ImageNet-R using \textbf{ViT-B/16}~\cite{dosovitskiy2021image} supervisedly pre-trained on ImageNet-21K~\cite{ridnik2021imagenet21k}.}
    \resizebox{\textwidth}{!}{
    \begin{tabular}{lcc@{~~}c@{~~}c@{~~}c@{~~}c@{~~}c@{~~}c@{~~}c}
    
    \toprule
       \multirow{2}{*}{\textbf{Method}} &\textbf{Memory} & \multicolumn{2}{c}{\textbf{CUB-200}} & \multicolumn{2}{c}{\textbf{Cars-196}} & \multicolumn{2}{c}{\textbf{CIFAR-100}} & \multicolumn{2}{c}{\textbf{ImageNet-R}} \\
        & \textbf{-Free}& Last-Acc & Inc-Acc & Last-Acc & Inc-Acc & Last-Acc & Inc-Acc & Last-Acc& Inc-Acc \\ 
        \midrule
        Joint-Training & - & 88.00\tiny{$\pm 0.34$} & - & 80.31\tiny{$\pm 0.13$} & - & 93.22\tiny{$\pm 0.16$} & - & 79.60\tiny{$\pm 0.87$} & - \\ 
        \midrule

        Prototype-classifier~\cite{janson2023simple} & \checkmark  & 80.66\tiny{$\pm 0.00$} & 88.95\tiny{$\pm 0.00$} & 28.58\tiny{$\pm 0.00$} & 39.83\tiny{$\pm 0.00$} & 60.29\tiny{$\pm 0.00$} & 69.18\tiny{$\pm 0.00$} & 38.45\tiny{$\pm 0.00$} & 45.59\tiny{$\pm 0.00$} \\
        
        GDumb \cite{prabhu2020gdumb}& & 61.80\tiny{$\pm 0.77$} & 79.76\tiny{$\pm 0.18$} & 25.20\tiny{$\pm 0.84$} & 49.48\tiny{$\pm 0.74$} & 81.92\tiny{$\pm 0.15$} & 89.46\tiny{$\pm 0.94$} & 24.23\tiny{$\pm 0.35$} & 43.48\tiny{$\pm 0.49$} \\
        
        DER++ \cite{buzzega2020dark}&  & 77.42\tiny{$\pm 0.71$} & 87.61\tiny{$\pm 0.09$} & 60.41\tiny{$\pm 1.76$} & 75.04\tiny{$\pm 0.57$} & 84.50\tiny{$\pm 1.67$} & 91.49\tiny{$\pm 0.61$} & 67.75\tiny{$\pm 0.93$} & 78.13\tiny{$\pm 1.14$} \\
        
        BiC \cite{wu2019large}& & 81.91\tiny{$\pm 2.59$} & 89.29\tiny{$\pm 1.57$} & 63.10\tiny{$\pm 5.71$} & 73.75\tiny{$\pm 2.37$} & 88.45\tiny{$\pm 0.57$} & 93.37\tiny{$\pm 0.32$} & 64.89\tiny{$\pm 0.80$} & 73.66\tiny{$\pm 1.61$} \\ 
        
        L2P \cite{wang2022learning}& $\checkmark$ & 62.21\tiny{$\pm 1.92$} & 73.83\tiny{$\pm 1.67$} & 38.18\tiny{$\pm 2.33$} & 51.79\tiny{$\pm 4.19$} & 82.76\tiny{$\pm 1.17$} & 88.48\tiny{$\pm 0.83$} & 66.49\tiny{$\pm 0.40$} & 72.83\tiny{$\pm 0.56$} \\ 
        
        DualPrompt \cite{wang2022dualprompt}& $\checkmark$ & 66.00\tiny{$\pm 0.57$} & 77.92\tiny{$\pm 0.50$} & 40.14\tiny{$\pm 2.36$} & 56.74\tiny{$\pm 1.78$} & 85.56\tiny{$\pm 0.33$} & 90.33\tiny{$\pm 0.33$} & 68.50\tiny{$\pm 0.52$} & 72.59\tiny{$\pm 0.24$} \\ 
        
        EWC \cite{kirkpatrick2017overcoming}& $\checkmark$ & 68.32\tiny{$\pm 2.64$} & 79.95\tiny{$\pm 2.28$} & 52.50\tiny{$\pm 3.18$} & 64.01\tiny{$\pm 3.25$} & 89.30\tiny{$\pm 0.23$} & 92.31\tiny{$\pm 1.66$} & 70.27\tiny{$\pm 1.99$} & 76.27\tiny{$\pm 2.13$} \\

        LwF \cite{li2017learning}& $\checkmark$ & 69.75\tiny{$\pm 1.37$} & 80.45\tiny{$\pm 2.08$} & 49.94\tiny{$\pm 3.24$} & 63.28\tiny{$\pm 1.11$} & 87.99\tiny{$\pm 0.05$} & 92.13\tiny{$\pm 1.16$} & 67.29\tiny{$\pm 1.67$} & 74.47\tiny{$\pm 1.48$} \\
        
        Seq FT & $\checkmark$ & 68.07\tiny{$\pm 1.09$} & 79.04\tiny{$\pm 1.69$} & 49.74\tiny{$\pm 1.25$} & 62.83\tiny{$\pm 2.16$} & 88.86\tiny{$\pm 0.83$} & 92.01\tiny{$\pm 1.71$} & 71.80\tiny{$\pm 1.45$} & 76.84\tiny{$\pm 1.26$} \\

        RanPAC~\cite{mcdonnell2024ranpac} & \checkmark & 85.82\tiny{$\pm 0.53$} & 91.47\tiny{$\pm 0.96$} & 53.84\tiny{$\pm 0.84$} & 66.39\tiny{$\pm 1.18$} & 90.09\tiny{$\pm 0.25$} & 93.31\tiny{$\pm 0.98$} & 72.62\tiny{$\pm 0.11$} & 78.35\tiny{$\pm 0.58$} \\
        
        SLCA~\cite{SLCA} & $\checkmark$ & 84.71\tiny{$\pm 0.40$} & 90.94\tiny{$\pm 0.68$} & 67.73\tiny{$\pm 0.85$} & 76.93\tiny{$\pm 1.21$} & 91.53\tiny{$\pm 0.28$} & 94.09\tiny{$\pm 0.87$} & 77.00\tiny{$\pm 0.33$} & 81.17\tiny{$\pm 0.64$} \\

        \midrule
        \rowcolor{LightCyan}
        \multirow{1}*{{CoMA (Ours)}} & \checkmark
        & 85.95\tiny{$\pm0.29$} & 90.75\tiny{$\pm0.39$} 
        & 73.35\tiny{$\pm0.59$}& 78.55\tiny{$\pm0.42$} 
        & 92.00\tiny{$\pm 0.13$} & 94.12\tiny{$\pm0.63$} 
        & 77.47\tiny{$\pm0.05$} & 81.32\tiny{$\pm0.17$} \\

        \rowcolor{LightCyan}
        CoFiMA (Ours) & \checkmark & 
        \textbf{87.11}\tiny{$\pm0.56$} & \textbf{91.87}\tiny{$\pm0.69$} & \textbf{76.96}\tiny{$\pm0.64$} & \textbf{82.65}\tiny{$\pm0.96$} & \textbf{92.77}\tiny{$\pm 0.24$} & \textbf{94.89}\tiny{$\pm 0.94$} & \textbf{78.25}\tiny{$\pm 0.26$} &\textbf{81.48}\tiny{$\pm0.56$} \\
        \bottomrule
    \end{tabular}
    }
    \label{tab.main1}
\end{table}
\subsection{State-of-the-art Comparison}
\label{main-analysis}
This section analyzes the performance of CoMA and CoFiMA across various CL benchmarks. We report in Tabs.~\ref{tab.main1}~and~\ref{tab.main2} the comparison of CoMA and CoFiMA with state-of-the-art CL methods using the ViT-B/16 backbone that was pre-trained supervisedly and unsupervisedly, respectively.

As shown in Tab.~\ref{tab.main1}, our proposed CoMA consistently outperforms the best-performing CIL baselines, SLCA~\cite{SLCA} and RanPAC~\cite{mcdonnell2024ranpac}, across all benchmarks. This confirms the benefits of model averaging in CIL. Moreover, CoFiMA, the improved variant of CoMA, shows even further performance improvement over CoMA, achieving new state-of-the-art results in CIL. This highlights the need for adaptive model averaging based on the importance of the parameters for a given task. In detail, \method achieves a Last-Acc of 87.11\% and an Inc-Acc of 91.87\% on CUB-200. This performance surpasses SLCA's performance, demonstrating a gain of \textbf{+2.4\%} and \textbf{+0.93\%} in Last-Acc and Inc-Acc, respectively. Similarly, on Cars-196, \method outperforms SLCA with \textbf{+9.23\%} and \textbf{+5.72\%} improvements in Last-Acc and Inc-Acc, respectively. This trend of outperforming SLCA is also observed in Imagenet-R, with \textbf{+1.25\%} and \textbf{+0.31\%} improvements in Last-Acc and Inc-Acc over SLCA, respectively.

\begin{table}[t]
    \centering
    \caption{State-of-the-art comparison on CUB-200, Cars-196, CIFAR-100, and ImageNet-R using \textbf{ViT-B/16} with self-supervised pre-training (\textbf{MoCo-V3}~\cite{chen2021mocov3}) on ImageNet-1K.}
   \resizebox{\textwidth}{!}{
    \begin{tabular}{lcc@{~~}c@{~~}c@{~~}c@{~~}c@{~~}c@{~~}c@{~~}c}
    \toprule
        \multirow{2}{*}{\textbf{Method}} & \textbf{Memory} & \multicolumn{2}{c}{\textbf{CUB-200}} & \multicolumn{2}{c}{\textbf{Cars-196}} & \multicolumn{2}{c}{\textbf{CIFAR-100}} & \multicolumn{2}{c}{\textbf{ImageNet-R}} \\ 
        & \textbf{-Free} & Last-Acc & Inc-Acc & Last-Acc & Inc-Acc & Last-Acc & Inc-Acc & Last-Acc & Inc-Acc \\ 
        \midrule
        Joint-Training & - & 79.55\tiny{$\pm 0.04$} & - & 74.52\tiny{$\pm 0.09$} & - & 89.11\tiny{$\pm 0.06$} & - & 72.80\tiny{$\pm 0.23$} & -  \\ 
        \midrule 

        Prototype-classifier~\cite{janson2023simple} & \checkmark & 51.57\tiny{$\pm 0.00$} &  63.43\tiny{$\pm 0.00$} & 20.97\tiny{$\pm 0.00$} & 30.10\tiny{$\pm 0.00$} & 73.50\tiny{$\pm 0.00$} & 81.75\tiny{$\pm 0.00$} & 37.60\tiny{$\pm 0.00$} & 44.95\tiny{$\pm 0.00$} \\
        
        GDumb \cite{prabhu2020gdumb}&  & 45.29\tiny{$\pm 0.97$} & 66.86\tiny{$\pm 0.63$} & 20.95\tiny{$\pm 0.42$} & 45.40\tiny{$\pm 0.66$} & 69.72\tiny{$\pm 0.20$} & 80.95\tiny{$\pm 1.19$} & 28.24\tiny{$\pm 0.58$} & 43.64\tiny{$\pm 1.05$} \\ 
        
        DER++ \cite{buzzega2020dark}&  & 61.47\tiny{$\pm 0.32$} & 77.15\tiny{$\pm 0.61$} & 50.64\tiny{$\pm 0.70$} & 67.64\tiny{$\pm 0.45$} & 63.64\tiny{$\pm 1.30$} & 79.55\tiny{$\pm 0.87$} & 53.11\tiny{$\pm 0.44$} & 65.10\tiny{$\pm 0.91$} \\  
        
        BiC \cite{wu2019large}& & 74.39\tiny{$\pm 1.12$} & 82.13\tiny{$\pm 0.33$} & 65.57\tiny{$\pm 0.93$} & 73.95\tiny{$\pm 0.29$} & 80.57\tiny{$\pm 0.86$} & 89.39\tiny{$\pm 0.33$} & 57.36\tiny{$\pm 2.68$} & 68.07\tiny{$\pm 0.22$} \\  
        
        EWC \cite{kirkpatrick2017overcoming}& $\checkmark$  & 61.36\tiny{$\pm 1.43$} & 72.84\tiny{$\pm 2.18$}  & 53.16\tiny{$\pm 1.45$} & 63.61\tiny{$\pm 1.06$} & 81.62\tiny{$\pm 0.34$} & 87.56\tiny{$\pm 0.97$} & 64.50\tiny{$\pm 0.36$} & 70.37\tiny{$\pm 0.41$} \\ 
        
        LwF \cite{li2017learning}& $\checkmark$ & 61.66\tiny{$\pm 1.95$} & 73.90\tiny{$\pm 1.91$}  & 52.45\tiny{$\pm 0.48$} & 63.87\tiny{$\pm 0.31$}  & 77.94\tiny{$\pm 1.00$} & 86.90\tiny{$\pm 0.90$} & 60.74\tiny{$\pm 0.30$} & 68.55\tiny{$\pm 0.65$}  \\ 
        
        Seq FT & $\checkmark$ & 61.67\tiny{$\pm 1.37$} & 73.25\tiny{$\pm 1.83$} & 52.91\tiny{$\pm 1.61$} & 63.32\tiny{$\pm 1.31$} & 81.47\tiny{$\pm 0.55$} & 87.55\tiny{$\pm 0.95$} & 64.43\tiny{$\pm 0.44$} & 70.48\tiny{$\pm 0.54$} \\

        RanPAC~\cite{mcdonnell2024ranpac} & $\checkmark$ & 74.43\tiny{$\pm 0.43$} & 83.63\tiny{$\pm 0.01$} & 63.21\tiny{$\pm 0.02$} & 74.01\tiny{$\pm 0.47$} & 86.47\tiny{$\pm 0.52$} & 90.81\tiny{$\pm 1.05$} & 69.11\tiny{$\pm 0.69$} & 75.20\tiny{$\pm 0.34$} \\
        
        SLCA~\cite{SLCA}& $\checkmark$ & 73.01\tiny{$\pm 0.16$} & 82.13\tiny{$\pm 0.34$} & 66.04\tiny{$\pm 0.08$} & 72.59\tiny{$\pm 0.04$} & 85.27\tiny{$\pm 0.08$} & 89.51\tiny{$\pm 1.04$} & 68.07\tiny{$\pm 0.21$} & 73.04\tiny{$\pm 0.56$} \\ 
        
        \midrule
        \rowcolor{LightCyan}
        CoMA (Ours) & $\checkmark $
        & 75.12\tiny{$\pm 0.27$} & 82.76\tiny{$\pm 0.16$} & 67.48\tiny{$\pm 0.19$} & 74.90\tiny{$\pm 0.87$} & 86.59\tiny{$\pm 0.51$} & 91.02\tiny{$\pm 0.47$} & 69.33\tiny{$\pm 0.22$} & 75.64\tiny{$\pm 0.13$} \\

        \rowcolor{LightCyan}
        CoFiMA (Ours) & \checkmark & 
        \textbf{77.65}\tiny{$\pm 0.18$} & \textbf{83.54}\tiny{$\pm 0.16$} & \textbf{69.51}\tiny{$\pm 0.16$} & \textbf{76.21}\tiny{$\pm 0.83$} & \textbf{87.44}\tiny{$\pm 0.47$} & \textbf{91.13}\tiny{$\pm 0.53$} & \textbf{70.87}\tiny{$\pm 0.31$} & \textbf{76.09}\tiny{$\pm 0.78$} \\
        \bottomrule
    \end{tabular}
    }
\label{tab.main2}
\end{table}

From Tab.~\ref{tab.main2} we observe that CoMA and CoFiMA both surpass the state-of-the-art methods by substantial margins while using a PTM pre-trained unsupervisedly on ImageNet-1K. The results confirm that having access to an unsupervised PTM is sufficient to reach satisfactory performance in CIL, although the absolute performance is lower for all methods compared to Tab.~\ref{tab.main1}. In detail, on CUB-200, CoFiMA's Last-Acc of 77.65\% and Inc-Acc of 83.54\% continue to show an advantage over SLCA. CoFiMA also leads in Cars-196, CIFAR-100, and ImageNet-R, showcasing its consistent performance across different datasets.

Notably, \method's performance is not far from the joint-training baselines in both Tabs.~\ref{tab.main1}~and~\ref{tab.main2}. For instance, in Tab.~\ref{tab.main1} for CIFAR-100, \method's Last-Acc is only \textbf{0.45\%} lower than the joint-training baseline of 93.22\%. This gap narrows further in other benchmarks, indicating CoFiMA's effectiveness in approaching the upper bounds of CL performance. \method, which efficiently balances the retention of old knowledge with the acquisition of new information, contributes to its strong performance using both supervised and unsupervised PTMs.

\subsection{Ablation Studies}
\label{sec.ablation}

\noindent \textbf{Analysis of Model Averaging.} This section evaluates our continual model-averaging approach against two baselines:  
\begin{itemize}
    \item \textit{\textbf{Weight-Ensemble}}, which uniformly averages model weights (\eg $\lambda\!=\!1/t$), initializing each model $M_t$ from either the pre-trained PTM ($\thetavect_0$) or the previous task's parameters ($\thetavect_{t-1}$). 
    \item \textit{\textbf{Exponential Moving Average (EMA)}}~\cite{szegedy2015rethinking}, a technique for a running average of model parameters computed at every gradient descent iteration $m$ as follows: $\thetavect_{m}\!=\!\beta \thetavect_{m} + (1 - \beta) \thetavect_{m\!-\!1}$ with $\beta=0.999$. 
\end{itemize}

\begin{table}[t]
 \centering
 \caption{Experimental results comparing our methods (CoMA, and \method) to weight-averaging baselines using \textbf{ViT-B/16}~\cite{dosovitskiy2021image} supervised PTM.} 
 \resizebox{\textwidth}{!}{
	\begin{tabular}{lccc@{~~}c@{~~}c@{~~}c@{~~}c@{~~}c@{~~}c@{~~}c@{~~}}
	 \toprule
  \textbf{Variant} &    \textbf{Init.} &$ \boldsymbol{\lambda}$& \multicolumn{2}{c}{\textbf{CUB-200}} &
  \multicolumn{2}{c}{\textbf{Cars-196}} & 
  \multicolumn{2}{c}{\textbf{CIFAR-100}} &
  \multicolumn{2}{c}{\textbf{Imagenet-R}} \\
   &&& Last-Acc & Inc-Acc & Last-Acc& Inc-Acc        & Last-Acc & Inc-Acc & Last-Acc & Inc-Acc \\
        \midrule
        Prototype-classifier~\cite{janson2023simple} & - & -   & 80.66\tiny{$\pm 0.00$} &  88.95\tiny{$\pm 0.00$} &28.58\tiny{$\pm 0.00$} & 39.83\tiny{$\pm 0.00$} & 60.29\tiny{$\pm 0.00$} & 69.18\tiny{$\pm 0.00$} & 38.45\tiny{$\pm 0.00$} & 45.59\tiny{$\pm 0.00$} \\ 
        
        Seq FT & - & -  &
        68.07\tiny{$\pm 1.09$} & 79.04\tiny{$\pm 1.69$} & 49.74\tiny{$\pm 1.25$} & 62.83\tiny{$\pm 2.16$} & 88.86\tiny{$\pm 0.83$} & 92.01\tiny{$\pm 1.71$} & 71.80\tiny{$\pm 1.45$} & 76.84\tiny{$\pm 1.26$}  \\
       
        SLCA~\cite{SLCA} & -& - & 
        84.71\tiny{$\pm 0.40$} & 90.94\tiny{$\pm 0.68$} & 67.73\tiny{$\pm 0.85$} & 76.93\tiny{$\pm 1.21$} & 91.53\tiny{$\pm 0.28$} & 94.09\tiny{$\pm 0.87$} & 77.00\tiny{$\pm 0.33$} & 81.17\tiny{$\pm 0.64$}  \\
       
        \midrule
        Weight-Ens.&$\theta_0$&$1/t$&
         82.49\tiny{$\pm0.27$} & 87.20\tiny{$\pm0.56$}
         & 36.20\tiny{$\pm0.27$} & 45.31\tiny{$\pm0.31$} &61.68\tiny{$\pm0.14$} & 70.24\tiny{$\pm0.46$} &44.90\tiny{$\pm0.14$} & 52.03\tiny{$\pm0.46$} \\
         
        Weight-Ens.&$\theta^*_{t-1}$&$1/t$ &
        84.28\tiny{$\pm0.47$} & 90.07\tiny{$\pm 0.22$}  
        & 71.82\tiny{$\pm 0.47$} & 78.85\tiny{$\pm 0.31$} 
        &91.69\tiny{$\pm0.23$} & 94.52\tiny{$\pm 0.95$}
        & 75.85\tiny{$\pm0.73$} & 81.51\tiny{$\pm 0.60$} \\ 

        \multirow{1}*{{EMA}} &-&-
        & 84.99\tiny{$\pm0.49$} & 90.84\tiny{$\pm0.74$}
        & 65.41\tiny{$\pm0.18$} & 73.68\tiny{$\pm0.46$}
        & 91.40\tiny{$\pm0.12$} & 93.89\tiny{$\pm0.59$} 
        & 77.35\tiny{$\pm0.83$} & 83.07\tiny{$\pm0.94$} \\
        
        \cdashline{1-11}
        \rowcolor{LightCyan}
        \multirow{1}*{{CoMA (Ours)}}& $\theta^*_{t-1}$ & $\lambda$
        & 85.95\tiny{$\pm0.29$} & 90.75\tiny{$\pm0.39$} 
        & 73.35\tiny{$\pm0.59$}& 78.55\tiny{$\pm0.42$} 
        & 92.00\tiny{$\pm 0.13$} & 94.12\tiny{$\pm0.63$} 
        & 77.47\tiny{$\pm0.05$} & 81.32\tiny{$\pm0.17$} \\
        \rowcolor{LightCyan}
        \multirow{1}*{{CoFiMA (Ours)}}& $\theta^*_{t-1}$ & $\lambda F_t$ &
        \textbf{87.11}\tiny{$\pm0.56$} & \textbf{91.87}\tiny{$\pm0.69$} & \textbf{76.96}\tiny{$\pm0.64$} & \textbf{82.65}\tiny{$\pm0.96$} & \textbf{92.77}\tiny{$\pm 0.24$} & \textbf{94.89}\tiny{$\pm 0.94$} & \textbf{78.25}\tiny{$\pm 0.26$} &\textbf{81.48}\tiny{$\pm0.56$} \\
       \bottomrule
	\end{tabular}
    }
 \label{tab:ablation_c}
\end{table}

Tab.~\ref{tab:ablation_c} presents the results. \method shows superior performance across all datasets. The \textit{Weight-Ensemble} with $\theta_{t-1}$ initialization yields competitive results in CIFAR-100 (Last-Acc: 91.69\%) but underperforms in datasets like Cars-196 (Last-Acc: 71.82\%). The \textit{Weight-Ensemble} starting from $\theta_0$ shows lower performance. This indicates the limitations of reinitialization to $\theta_0$; as models are trained on different tasks, it leads to different optima, as depicted in Fig.~\ref{fig:pipeline}(a). 

However, \textit{Weight-Ensemble} with $\thetavect_{t\!-\!1}$ initialization outperforms the variant starting from $\thetavect_0$, emphasizing the importance of initialization. Initializing $\thetavect_{t}$ from $\thetavect^{*}_{t\!-\!1}$ improves performance on both task $t$ and $t\!-\!1$, as done in~\cite{wortsman2021robust, ilharco2022patching}. This gain in performance indicates that averaging successive models leads to good performance on the current task $t$ while preserving previous knowledge. 

Averaging all models trained up to task $t$ is suboptimal, as weights at task $t$ likely differ significantly from those at task $t\!=\!1$. Such averaging shifts the $\thetavect^{*}_{t}$ values toward suboptimal minima, leading to decreased performance (refer to Fig.~\ref{fig:pipeline}(b)). In contrast, \method avoids this by averaging only with the preceding parameters $\thetavect^{*}_{t-1}$, leading to better performance.

The \textit{EMA} method, though superior to both \textit{Weight-Ensemble} variants in most scenarios, falls short of \method's performance. This discrepancy may stem from excessive averaging in \textit{EMA}, where $\thetavect^{*}_{t}$ is modified multiple times within the same task, potentially leading to suboptimal outcomes. Differently, our method applies weight averaging only after completing each task, enhancing computational efficiency by selectively focusing on pertinent model weights while retaining previous knowledge.

\begin{figure}[t]
    \centering 
    \begin{minipage}{0.49\textwidth}
        \centering 
        \includegraphics[trim={0 20 0 20}, clip, width=\linewidth]{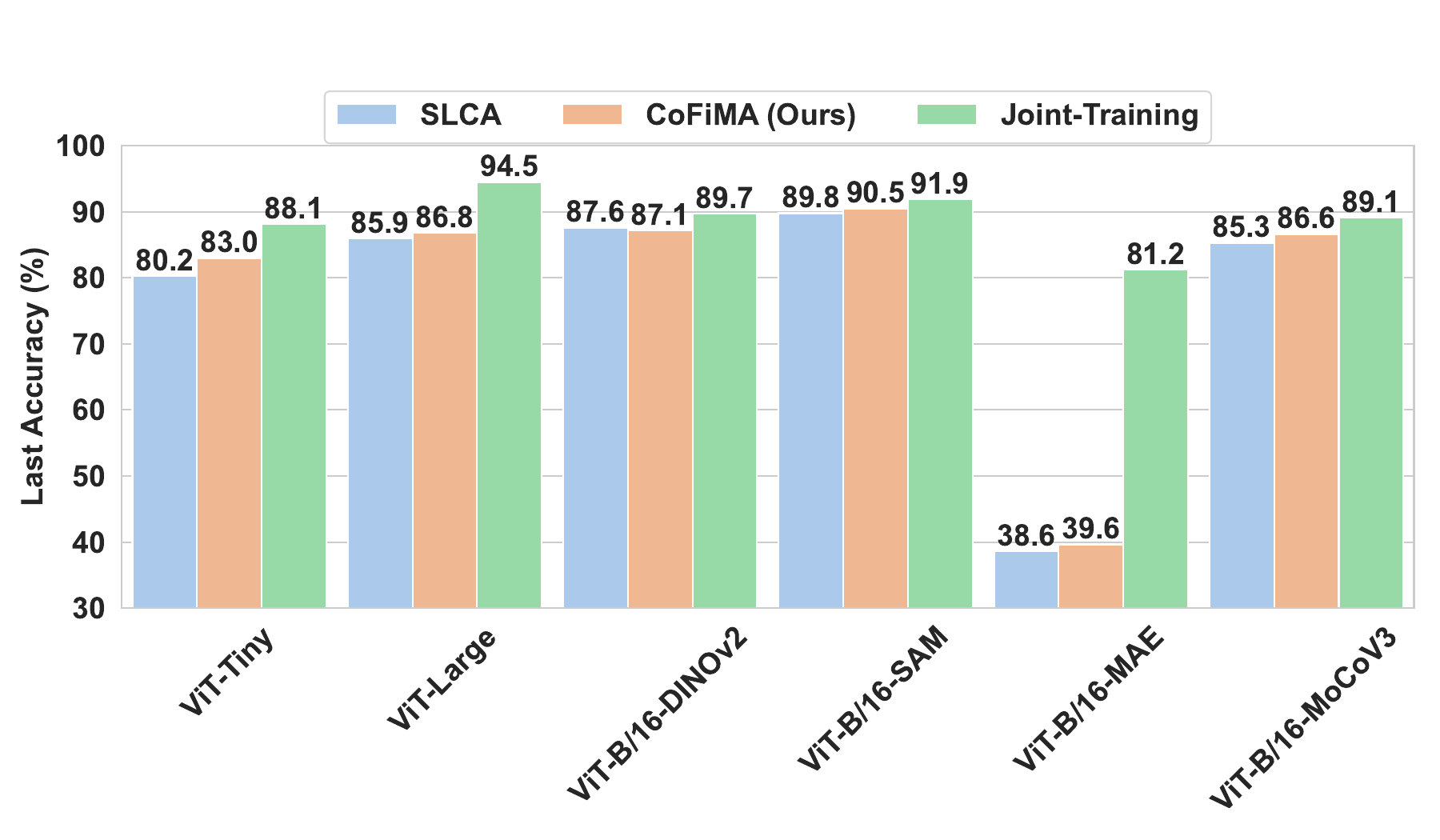}
        \caption{\method with various PTMs on CIFAR-100. \method enhances the results of SLCA. }
        \label{fig:ptms_cifar}
    \end{minipage}\hfill
    \begin{minipage}{0.49\textwidth}
        \centering 
        \includegraphics[trim={0 8 0 10}, clip, width=\linewidth]{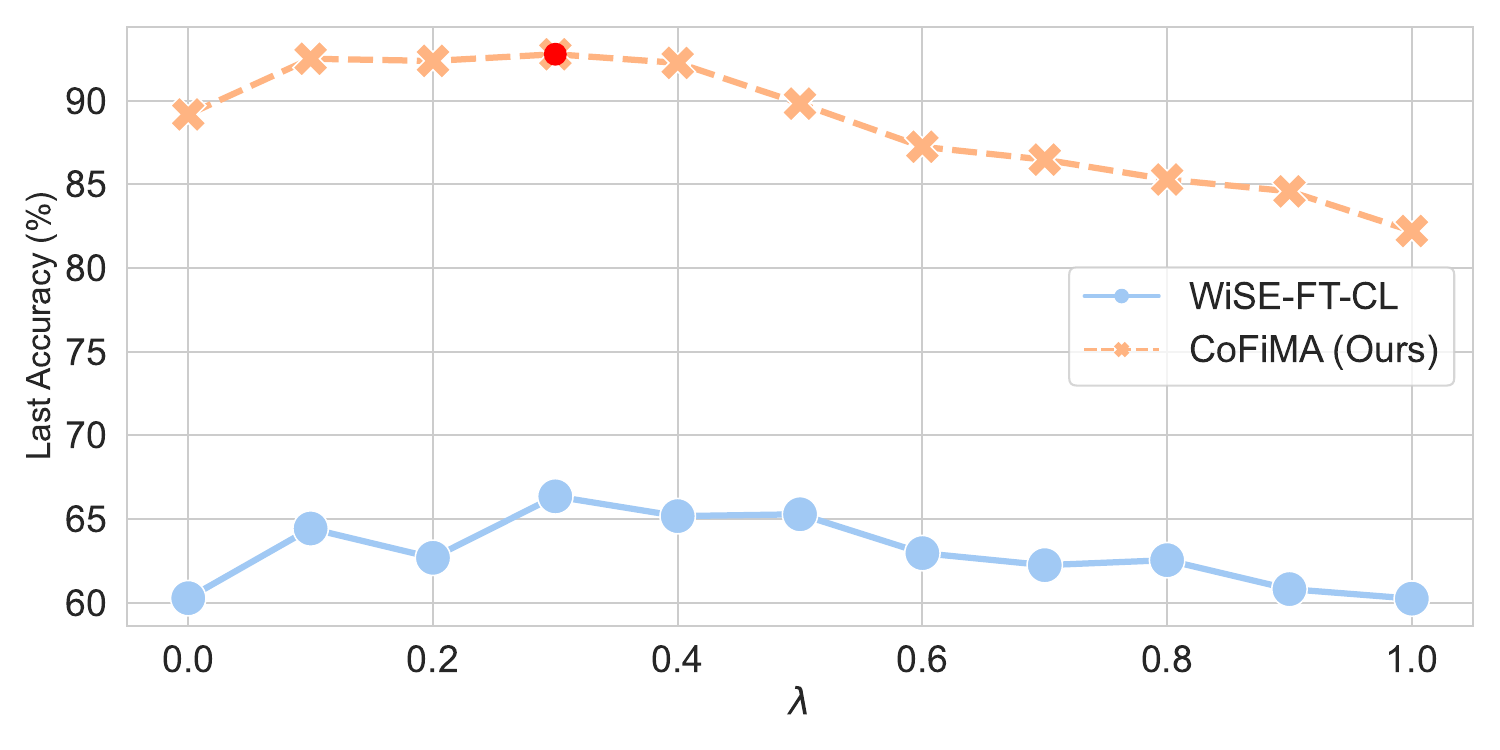}
        \caption{Ablation study on the effect of $\lambda$ on CIFAR-100 dataset. The red marker point represents the best performance.}
        \label{fig_lambda}
    \end{minipage}
\end{figure}

\noindent \textbf{Effect of PTMs.} In this section, we evaluate the performance of the \method approach across a variety of backbone architectures, including self-supervised (MAE~\cite{he2021masked}, MoCoV3~\cite{chen2021mocov3}, and DINOv2~\cite{dinoV2}) and supervised (ViT-Tiny~\cite{dosovitskiy2021image} and ViT-B/16-SAM~\cite{foret2021sharpnessaware}) models. This comprehensive analysis aims to ascertain the adaptability and performance consistency of CoFiMA in diverse training paradigms. Results are visualized in Fig.~\ref{fig:ptms_cifar}.

Our results indicate that \method enhances performance across almost all tested backbones relative to the baseline SLCA method. For instance, with the ViT-Tiny backbone, CoFiMA improves the Last-Accuracy of SLCA from 80.25\% to 82.96\%. This trend is similarly observed with ViT-Large, where CoFiMA achieves an accuracy of 86.81\%, surpassing SLCA's 85.93\%. Only in the case of ViT-B/16-DINOv2, there is a slight reduction in performance, likely due to the benchmark reaching its saturation point: both SLCA and \method exhibit performance akin to the joint-learning gold standard.

In the context of self-supervised learning models, \method demonstrates its efficacy by outperforming SLCA on ViT-B/16-MAE and ViT-B/16-MoCoV3 backbones. Importantly, the ViT-B/16-SAM backbone~\cite{foret2021sharpnessaware} achieves the best performance among the evaluated models. This can be attributed to the effective generalization features ingrained in the SAM backbone, a result of being trained with the SAM optimizer. This optimizer is known for its capability to enhance model generalizability, which is reflected in the superior performance metrics observed in our experiments, as also mentioned in Mehta~\etal~\cite{mehta2023empirical} work. We also notice that self-supervised pre-training usually results in larger performance gaps between continual learning baselines and joint training. Especially for ViT-B/16-MAE, as noted in Zhang~\etal~\cite{SLCA} work, because joint training using MAE requires smaller updates to learn all tasks compared to incremental learning using SLCA or CoFiMA.

When compared to joint training, a standard upper bound in CL settings, \method demonstrates competitive performance. Although joint training always leads, as seen with ViT-Large (94.45\%) and ViT-B/16-SAM (91.87\%), \method remains close, particularly with ViT-B/16-SAM, where it achieves 90.48\%. These results indicate that \method is a versatile approach, effective in enhancing performance with various backbones in both supervised and self-supervised learning contexts. However, the performance boost from our approach varies with the choice of backbone (size) and its pre-training paradigm.

\noindent\textbf{Balancing the Information of an Old and a New Task. }In CL, a primary objective is to balance knowledge retention from previous tasks with the acquisition of new information from current tasks. This study examines the impact of $\lambda$ used to balance the stability/plasticity of the model during averaging.

In addition to our method, we also include an adaptation of the aggregation scheme of WiSE-FT~\cite{wortsman2021robust} for CL, which we refer to as \textbf{WiSE-FT-CL}. After learning task $t$, stability is achieved via linear interpolation between the pre-trained and the fine-tuned model: $\thetavect_{t}^{*} = \lambda \thetavect_{t} + (1 - \lambda) \thetavect_{0}$. For every new task $t$, fine-tuning starts from $\thetavect_{t-1}^{*}$. This method aligns with WiSE-FT~\cite{wortsman2021robust} and relies only on the initial pre-training to achieve stability. 

According to Fig.~\ref{fig_lambda}, CoFiMA demonstrates superior performance compared to WiSE-FT-CL across various $\lambda$ settings. This improvement suggests that our method is more effective at incorporating new knowledge while preserving old information. The best performance is achieved with $\lambda\!=\!0.3$, giving the best trade-off between learning new task $t$ (\ie, plasticity) and preserving knowledge from previous tasks (\ie, stability). 

The underperformance of WiSE-FT-CL is attributed to the impact of averaging with the model $\thetavect_0$, which possesses significantly different parameter values than $\thetavect_t$ due to fine-tuning. Consequently, this averaging process results in suboptimal parameter values as explained in Sec.~\ref{method}. For CoFiMA, higher values for $\lambda$ put more emphasis on the current model at task $t$, thus leading to forgetting previous tasks. 

In summary, our method, CoFiMA, effectively maintains a balance between retaining old task information and adapting to new task data. This is achieved by leveraging weight averaging between successive models. The key benefit here is that the parameter values of models at tasks $t$ and $t-1$ do not diverge significantly, ensuring that the averaged parameters remain effective for both tasks~\cite{chizat2020lazy, ilharco2022patching} (more details in supplementary material).

\section{Conclusion}
In this work, we have introduced CoFiMA, the first method based on weight averaging techniques to address catastrophic forgetting in the CIL setting. This approach builds its grounding on two pillars. First, it leverages model averaging, providing a balanced mechanism for retaining prior knowledge while accommodating new information. Second, Fisher information is incorporated to intelligently weigh the averaging of parameters. This refinement allows for the adjustment of each parameter's value based on its importance, as determined by its Fisher information, thereby effectively reducing catastrophic forgetting. 

Our benchmarking on diverse datasets with various PTM backbones demonstrates that CoFiMA consistently outperforms state-of-the-art CIL methods. Our findings underscore the efficacy of CoFiMA in mitigating forgetting and highlight its versatility across different PTM backbones and benchmark datasets.

\section*{Acknowledgements}
We thank Yasser Benigmim, Khalid Oublal and Ekaterina Iakovleva for helpful discussions and draft feedback. This paper has been supported by the French National Research Agency (ANR) in the framework of its JCJC, and was funded by the European Union's Horizon Europe research and innovation program under grant agreement No. 101120237 (ELIAS). This research was additionally funded by the EU project AI4TRUST (No.101070190). Furthermore, this research was partially funded by Hi!PARIS Center on Data Analytics and Artificial Intelligence. This work was granted access to the HPC resources of IDRIS under the allocation AD011013860 made
by GENCI.


%
%
\bibliographystyle{splncs04}
\bibliography{main}

\begin{thebibliography}{10}
\providecommand{\url}[1]{\texttt{#1}}
\providecommand{\urlprefix}{URL }
\providecommand{\doi}[1]{https://doi.org/#1}

\bibitem{abraham2005memory}
Abraham, W.C., Robins, A.: Memory retention--the synaptic stability versus plasticity dilemma. Trends in neurosciences  \textbf{28}(2),  73--78 (2005)

\bibitem{aljundi2018memory}
Aljundi, R., Babiloni, F., Elhoseiny, M., Rohrbach, M., Tuytelaars, T.: Memory aware synapses: Learning what (not) to forget. In: Eur. Conf. Comput. Vis. (2018)

\bibitem{NIPS1996_39e4973b}
Amari, S.i.: Neural learning in structured parameter spaces - natural riemannian gradient. In: Mozer, M., Jordan, M., Petsche, T. (eds.) Advances in Neural Information Processing Systems. vol.~9. MIT Press (1996)

\bibitem{buzzega2020dark}
Buzzega, P., Boschini, M., Porrello, A., Abati, D., Calderara, S.: Dark experience for general continual learning: a strong, simple baseline. Adv. Neural Inform. Process. Syst.  \textbf{33} (2020)

\bibitem{chen2021mocov3}
Chen*, X., Xie*, S., He, K.: An empirical study of training self-supervised vision transformers. arXiv preprint arXiv:2104.02057  (2021)

\bibitem{chizat2020lazy}
Chizat, L., Oyallon, E., Bach, F.: On lazy training in differentiable programming (2020)

\bibitem{choshen2022fusing}
Choshen, L., Venezian, E., Slonim, N., Katz, Y.: Fusing finetuned models for better pretraining (2022)

\bibitem{dhar2019learning}
Dhar, P., Singh, R.V., Peng, K.C., Wu, Z., Chellappa, R.: Learning without memorizing. In: IEEE Conf. Comput. Vis. Pattern Recog. (2019)

\bibitem{bayesrules}
Dogucu, M., Johnson, A., Ott, M.: bayesrules: Datasets and Supplemental Functions from Bayes Rules! Book (2021), r package version 0.0.2.9000

\bibitem{effect_scale_ptms}
Dyer, E., Lewkowycz, A., Ramasesh, V.: Effect of scale on catastrophic forgetting in neural networks. In: Int. Conf. Learn. Represent. (2022)

\bibitem{d2021convit}
d’Ascoli, S., Touvron, H., Leavitt, M.L., Morcos, A.S., Biroli, G., Sagun, L.: Convit: Improving vision transformers with soft convolutional inductive biases. In: International Conference of Machine Learning (2021)

\bibitem{entezari2021role}
Entezari, R., Sedghi, H., Saukh, O., Neyshabur, B.: The role of permutation invariance in linear mode connectivity of neural networks. In: Int. Conf. Learn. Represent. (2022)

\bibitem{foret2021sharpnessaware}
Foret, P., Kleiner, A., Mobahi, H., Neyshabur, B.: Sharpness-aware minimization for efficiently improving generalization. In: International Conference on Learning Representations (2021)

\bibitem{frankle2020linear}
Frankle, J., Dziugaite, G.K., Roy, D., Carbin, M.: Linear mode connectivity and the lottery ticket hypothesis. In: Int. Conf. Machine Learn. (2020)

\bibitem{french1999catastrophic}
French, R.M.: Catastrophic forgetting in connectionist networks. Trends in cognitive sciences  \textbf{3}(4) (1999)

\bibitem{friedman2001elements}
Friedman, J., Hastie, T., Tibshirani, R., et~al.: The elements of statistical learning. Springer series in statistics New York (2001)

\bibitem{he2021masked}
He, K., Chen, X., Xie, S., Li, Y., Dollár, P., Girshick, R.: Masked autoencoders are scalable vision learners (2021)

\bibitem{he2015residual}
He, K., Zhang, X., Ren, S., Sun, J.: Deep residual learning for image recognition. In: IEEE Conf. Comput. Vis. Pattern Recog. (2015)

\bibitem{hendrycks2021many}
Hendrycks, D., Basart, S., Mu, N., Kadavath, S., Wang, F., Dorundo, E., Desai, R., Zhu, T., Parajuli, S., Guo, M., et~al.: The many faces of robustness: A critical analysis of out-of-distribution generalization. In: International Conference of Computer Vision (ICCV) (2021)

\bibitem{hinton2015distilling}
Hinton, G., Vinyals, O., Dean, J.: Distilling the knowledge in a neural network. arXiv preprint arXiv:1503.02531  (2015)

\bibitem{ilharco2022patching}
Ilharco, G., Wortsman, M., Gadre, S.Y., Song, S., Hajishirzi, H., Kornblith, S., Farhadi, A., Schmidt, L.: Patching open-vocabulary models by interpolating weights (2022)

\bibitem{izmailov2018averaging}
Izmailov, P., Podoprikhin, D., Garipov, T., Vetrov, D., Wilson, A.G.: Averaging weights leads to wider optima and better generalization. In: Conference on Uncertainty in Artificial Intelligence (UAI) (2018)

\bibitem{janson2023simple}
Janson, P., Zhang, W., Aljundi, R., Elhoseiny, M.: A simple baseline that questions the use of pretrained-models in continual learning (2023)

\bibitem{kairouz2021advances}
Kairouz, P., McMahan, H.B., Avent, B., Bellet, A., Bennis, M., Bhagoji, A.N., Bonawitz, K., Charles, Z., Cormode, G., Cummings, R., D'Oliveira, R.G.L., Eichner, H., Rouayheb, S.E., Evans, D., Gardner, J., Garrett, Z., Gascón, A., Ghazi, B., Gibbons, P.B., Gruteser, M., Harchaoui, Z., He, C., He, L., Huo, Z., Hutchinson, B., Hsu, J., Jaggi, M., Javidi, T., Joshi, G., Khodak, M., Konečný, J., Korolova, A., Koushanfar, F., Koyejo, S., Lepoint, T., Liu, Y., Mittal, P., Mohri, M., Nock, R., Özgür, A., Pagh, R., Raykova, M., Qi, H., Ramage, D., Raskar, R., Song, D., Song, W., Stich, S.U., Sun, Z., Suresh, A.T., Tramèr, F., Vepakomma, P., Wang, J., Xiong, L., Xu, Z., Yang, Q., Yu, F.X., Yu, H., Zhao, S.: Advances and open problems in federated learning (2021)

\bibitem{Kim_2023_CVPR}
Kim, D., Han, B.: On the stability-plasticity dilemma of class-incremental learning. In: IEEE Conf. Comput. Vis. Pattern Recog. (2023)

\bibitem{kirkpatrick2017overcoming}
Kirkpatrick, J., Pascanu, R., Rabinowitz, N., Veness, J., Desjardins, G., Rusu, A.A., Milan, K., Quan, J., Ramalho, T., Grabska-Barwinska, A., et~al.: Overcoming catastrophic forgetting in neural networks. PNAS  \textbf{114}(13) (2017)

\bibitem{dosovitskiy2021image}
Kolesnikov, A., Dosovitskiy, A., Weissenborn, D., Heigold, G., Uszkoreit, J., Beyer, L., Minderer, M., Dehghani, M., Houlsby, N., Gelly, S., Unterthiner, T., Zhai, X.: An image is worth 16x16 words: Transformers for image recognition at scale. In: Int. Conf. Learn. Represent. (2021)

\bibitem{krause20133d}
Krause, J., Stark, M., Deng, J., Fei-Fei, L.: 3d object representations for fine-grained categorization. In: Proceedings of the IEEE International Conference on Computer Vision Workshops (2013)

\bibitem{stanfordcars}
Krause, J., Stark, M., Deng, J., Fei-Fei, L.: 3d object representations for fine-grained categorization. In: 2013 IEEE International Conference on Computer Vision Workshops (2013)

\bibitem{krizhevsky2009learning}
Krizhevsky, A., Hinton, G., et~al.: Learning multiple layers of features from tiny images (2009)

\bibitem{deepensembles}
Lakshminarayanan, B., Pritzel, A., Blundell, C.: Simple and scalable predictive uncertainty estimation using deep ensembles. In: Adv. Neural Inform. Process. Syst. (2017)

\bibitem{9084352}
Li, T., Sahu, A.K., Talwalkar, A., Smith, V.: Federated learning: Challenges, methods, and future directions. IEEE Signal Processing Magazine  \textbf{37}(3) (2020)

\bibitem{li2017learning}
Li, Z., Hoiem, D.: Learning without forgetting. TPAMI  \textbf{40}(12) (2017)

\bibitem{8493265}
Liao, Z., Drummond, T., Reid, I., Carneiro, G.: Approximate fisher information matrix to characterize the training of deep neural networks. IEEE Transactions on Pattern Analysis and Machine Intelligence  \textbf{42}(1) (2020)

\bibitem{liu2021swin}
Liu, Z., Lin, Y., Cao, Y., Hu, H., Wei, Y., Zhang, Z., Lin, S., Guo, B.: Swin transformer: Hierarchical vision transformer using shifted windows (2021)

\bibitem{masana2022class}
Masana, M., Liu, X., Twardowski, B., Menta, M., Bagdanov, A.D., van~de Weijer, J.: Class-incremental learning: survey and performance evaluation on image classification. IEEE Trans. Pattern Anal. Mach. Intell.  (2022)

\bibitem{matena2022merging}
Matena, M.S., Raffel, C.A.: Merging models with fisher-weighted averaging. Adv. Neural Inform. Process. Syst.  \textbf{35},  17703--17716 (2022)

\bibitem{mcdonnell2024ranpac}
McDonnell, M.D., Gong, D., Parvaneh, A., Abbasnejad, E., van~den Hengel, A.: Ranpac: Random projections and pre-trained models for continual learning. Adv. Neural Inform. Process. Syst.  \textbf{36} (2024)

\bibitem{mcmahan2023communicationefficient}
McMahan, H.B., Moore, E., Ramage, D., Hampson, S., y~Arcas, B.A.: Communication-efficient learning of deep networks from decentralized data (2023)

\bibitem{mehta2023empirical}
Mehta, S.V., Patil, D., Chandar, S., Strubell, E.: An empirical investigation of the role of pre-training in lifelong learning (2023)

\bibitem{mirzadeh2020linear}
Mirzadeh, S.I., Farajtabar, M., Gorur, D., Pascanu, R., Ghasemzadeh, H.: Linear mode connectivity in multitask and continual learning (2020)

\bibitem{murata2022learning}
Murata, K., Ito, S., Ohara, K.: Learning and transforming general representations to break down stability-plasticity dilemma. In: Proceedings of the Asian Conference on Computer Vision (2022)

\bibitem{neyshabur2021transferred}
Neyshabur, B., Sedghi, H., Zhang, C.: What is being transferred in transfer learning? In: Adv. Neural Inform. Process. Syst. (2020)

\bibitem{nichol2018first}
Nichol, A., Achiam, J., Schulman, J.: On first-order meta-learning algorithms. arXiv preprint arXiv:1803.02999  (2018)

\bibitem{dinoV2}
Oquab, M., Darcet, T., Moutakanni, T., Vo, H.V., Szafraniec, M., Khalidov, V., Fernandez, P., Haziza, D., Massa, F., El-Nouby, A., Howes, R., Huang, P.Y., Xu, H., Sharma, V., Li, S.W., Galuba, W., Rabbat, M., Assran, M., Ballas, N., Synnaeve, G., Misra, I., Jegou, H., Mairal, J., Labatut, P., Joulin, A., Bojanowski, P.: Dinov2: Learning robust visual features without supervision (2023)

\bibitem{ovadia2019can}
Ovadia, Y., Fertig, E., Ren, J., Nado, Z., Sculley, D., Nowozin, S., Dillon, J.V., Lakshminarayanan, B., Snoek, J.: Can you trust your model's uncertainty? evaluating predictive uncertainty under dataset shift. In: Adv. Neural Inform. Process. Syst. (2019)

\bibitem{panos2023first}
Panos, A., Kobe, Y., Reino, D.O., Aljundi, R., Turner, R.E.: First session adaptation: A strong replay-free baseline for class-incremental learning. In: Int. Conf. Comput. Vis. (2023)

\bibitem{pascanu2014revisiting}
Pascanu, R., Bengio, Y.: Revisiting natural gradient for deep networks (2014)

\bibitem{pham2021continual}
Pham, Q., Liu, C., Steven, H.: Continual normalization: Rethinking batch normalization for online continual learning. In: Int. Conf. Learn. Represent. (2022)

\bibitem{prabhu2020gdumb}
Prabhu, A., Torr, P.H., Dokania, P.K.: Gdumb: A simple approach that questions our progress in continual learning. In: Eur. Conf. Comput. Vis. (2020)

\bibitem{radford2021learning}
Radford, A., Kim, J.W., Hallacy, C., Ramesh, A., Goh, G., Agarwal, S., Sastry, G., Askell, A., Mishkin, P., Clark, J., et~al.: Learning transferable visual models from natural language supervision. In: Int. Conf. Machine Learn. PMLR (2021)

\bibitem{ramasesh2021effect}
Ramasesh, V.V., Lewkowycz, A., Dyer, E.: Effect of scale on catastrophic forgetting in neural networks. In: Proceedings of the International Conference on Learning Representations (2021)

\bibitem{ramé2023diverse}
Ramé, A., Kirchmeyer, M., Rahier, T., Rakotomamonjy, A., Gallinari, P., Cord, M.: Diverse weight averaging for out-of-distribution generalization (2023)

\bibitem{ren2024analyzing}
Ren, W., Li, X., Wang, L., Zhao, T., Qin, W.: Analyzing and reducing catastrophic forgetting in parameter efficient tuning (2024)

\bibitem{ridnik2021imagenet21k}
Ridnik, T., Ben-Baruch, E., Noy, A., Zelnik-Manor, L.: Imagenet-21k pretraining for the masses (2021)

\bibitem{rusu2016progressive}
Rusu, A.A., Rabinowitz, N.C., Desjardins, G., Soyer, H., Kirkpatrick, J., Kavukcuoglu, K., Pascanu, R., Hadsell, R.: Progressive neural networks. arXiv preprint arXiv:1606.04671  (2016)

\bibitem{divideforgetensembleselectively}
Rype{\'s}{\'c}, G., Cygert, S., Khan, V., Trzcinski, T., Zieli{\'n}ski, B.M., Twardowski, B.: Divide and not forget: Ensemble of selectively trained experts in continual learning. In: Int. Conf. Learn. Represent. (2023)

\bibitem{schuhmann2021laion}
Schuhmann, C., Vencu, R., Beaumont, R., Kaczmarczyk, R., Mullis, C., Katta, A., Coombes, T., Jitsev, J., Komatsuzaki, A.: Laion-400m: Open dataset of clip-filtered 400 million image-text pairs. arXiv preprint arXiv:2111.02114  (2021)

\bibitem{serra2018overcoming}
Serra, J., Suris, D., Miron, M., Karatzoglou, A.: Overcoming catastrophic forgetting with hard attention to the task. In: Int. Conf. Machine Learn. (2018)

\bibitem{simon2022generalizing}
Simon, C., Faraki, M., Tsai, Y.H., Yu, X., Schulter, S., Suh, Y., Harandi, M., Chandraker, M.: On generalizing beyond domains in cross-domain continual learning. arXiv preprint arXiv:2203.03970  (2022)

\bibitem{soen2021variance}
Soen, A., Sun, K.: On the variance of the fisher information for deep learning. Adv. Neural Inform. Process. Syst.  \textbf{34},  5708--5719 (2021)

\bibitem{spall2005monte}
Spall, J.C.: Monte carlo computation of the fisher information matrix in nonstandard settings. Journal of Computational and Graphical Statistics  \textbf{14}(4) (2005)

\bibitem{4586850}
Spall, J.C.: Improved methods for monte carlo estimation of the fisher information matrix. In: 2008 American Control Conference (2008)

\bibitem{stickland2020diverse}
Stickland, A.C., Murray, I.: Diverse ensembles improve calibration. In: International Conference on Machine Learning (ICML) Workshop on Uncertainty and Robustness in Deep Learning (2020)

\bibitem{sun2023pilot}
Sun, H.L., Zhou, D.W., Ye, H.J., Zhan, D.C.: Pilot: A pre-trained model-based continual learning toolbox. arXiv preprint arXiv:2309.07117  (2023)

\bibitem{szegedy2016rethinking}
Szegedy, C., Vanhoucke, V., Ioffe, S., Shlens, J., Wojna, Z.: Rethinking the inception architecture for computer vision. In: IEEE Conf. Comput. Vis. Pattern Recog. (2016)

\bibitem{szegedy2015rethinking}
Szegedy, C., Vanhoucke, V., Ioffe, S., Shlens, J., Wojna, Z.: Rethinking the inception architecture for computer vision. In: IEEE Conf. Comput. Vis. Pattern Recog. (2016)

\bibitem{titsias2020functional}
Titsias, M.K., Schwarz, J., de~G.~Matthews, A.G., Pascanu, R., Teh, Y.W.: Functional regularisation for continual learning with gaussian processes (2020)

\bibitem{vandeven2019scenarios}
van~de Ven, G.M., Tolias, A.S.: Three scenarios for continual learning (2019)

\bibitem{villa2022pivot}
Villa, A., Alc{\'a}zar, J.L., Alfarra, M., Alhamoud, K., Hurtado, J., Heilbron, F.C., Soto, A., Ghanem, B.: Pivot: Prompting for video continual learning. arXiv preprint arXiv:2212.04842  (2022)

\bibitem{WahCUB2002011}
Wah, C., Branson, S., Welinder, P., Perona, P., Belongie, S.: {The Caltech-UCSD Birds-200-2011 Dataset}. Tech. Rep. CNS-TR-2011-001, California Institute of Technology (2011)

\bibitem{wah2011caltech}
Wah, C., Branson, S., Welinder, P., et~al.: The caltech-ucsd birds-200-2011 dataset (2011)

\bibitem{wang2021ordisco}
Wang, L., Yang, K., Li, C., Hong, L., Li, Z., Zhu, J.: Ordisco: Effective and efficient usage of incremental unlabeled data for semi-supervised continual learning. In: IEEE Conf. Comput. Vis. Pattern Recog. (2021)

\bibitem{Wang_2023}
Wang, L., Zhang, X., Li, Q., Zhang, M., Su, H., Zhu, J., Zhong, Y.: Incorporating neuro-inspired adaptability for continual learning in artificial intelligence. Nature Machine Intelligence  \textbf{5}(12) (Nov 2023)

\bibitem{wang2022coscl}
Wang, L., Zhang, X., Li, Q., Zhu, J., Zhong, Y.: Coscl: Cooperation of small continual learners is stronger than a big one (2022)

\bibitem{wang2023comprehensive}
Wang, L., Zhang, X., Su, H., Zhu, J.: A comprehensive survey of continual learning: Theory, method and application (2023)

\bibitem{wang2022s}
Wang, Y., Huang, Z., Hong, X.: S-prompts learning with pre-trained transformers: An occam's razor for domain incremental learning. arXiv preprint arXiv:2207.12819  (2022)

\bibitem{wang2022dualprompt}
Wang, Z., Zhang, Z., Ebrahimi, S., Sun, R., Zhang, H., Lee, C.Y., Ren, X., Su, G., Perot, V., Dy, J., et~al.: Dualprompt: Complementary prompting for rehearsal-free continual learning. In: Eur. Conf. Comput. Vis. Springer (2022)

\bibitem{wang2022learning}
Wang, Z., Zhang, Z., Lee, C.Y., Zhang, H., Sun, R., Ren, X., Su, G., Perot, V., Dy, J., Pfister, T.: Learning to prompt for continual learning. In: IEEE Conf. Comput. Vis. Pattern Recog. (2022)

\bibitem{wortsman2022model}
Wortsman, M., Ilharco, G., Gadre, S.Y., Roelofs, R., Gontijo-Lopes, R., Morcos, A.S., Namkoong, H., Farhadi, A., Carmon, Y., Kornblith, S., et~al.: Model soups: averaging weights of multiple fine-tuned models improves accuracy without increasing inference time. In: Int. Conf. Machine Learn. PMLR (2022)

\bibitem{wortsman2021robust}
Wortsman, M., Ilharco, G., Li, M., Kim, J.W., Hajishirzi, H., Farhadi, A., Namkoong, H., Schmidt, L.: Robust fine-tuning of zero-shot models. In: IEEE Conf. Comput. Vis. Pattern Recog. (2022)

\bibitem{wu2022class}
Wu, T.Y., Swaminathan, G., Li, Z., Ravichandran, A., Vasconcelos, N., Bhotika, R., Soatto, S.: Class-incremental learning with strong pre-trained models. In: IEEE Conf. Comput. Vis. Pattern Recog. (2022)

\bibitem{wu2019large}
Wu, Y., Chen, Y., Wang, L., Ye, Y., Liu, Z., Guo, Y., Fu, Y.: Large scale incremental learning. In: IEEE Conf. Comput. Vis. Pattern Recog. (2019)

\bibitem{yang2022continual}
Yang, B., Deng, X., Shi, H., Li, C., Zhang, G., Xu, H., Zhao, S., Lin, L., Liang, X.: Continual object detection via prototypical task correlation guided gating mechanism. In: IEEE Conf. Comput. Vis. Pattern Recog. (2022)

\bibitem{zenke2017continual}
Zenke, F., Poole, B., Ganguli, S.: Continual learning through synaptic intelligence. In: Int. Conf. Machine Learn. (2017)

\bibitem{SLCA}
Zhang, G., Wang, L., Kang, G., Chen, L., Wei, Y.: Slca: Slow learner with classifier alignment for continual learning on a pre-trained model. In: Int. Conf. Comput. Vis. (2023)

\bibitem{Zhang_2020_WACV}
Zhang, J., Zhang, J., Ghosh, S., Li, D., Zhu, J., Zhang, H., Wang, Y.: Regularize, expand and compress: Nonexpansive continual learning. In: WACV (2020)

\bibitem{AdamAdapter}
Zhou, D.W., Ye, H.J., Zhan, D.C., Liu, Z.: Revisiting class-incremental learning with pre-trained models: Generalizability and adaptivity are all you need (2023)

\end{thebibliography}

\appendix
\clearpage
\setcounter{page}{1}

In this supplementary material, we provide more details about the experimental results mentioned in the main paper, as well as additional empirical evaluations and discussions. The supplementary material is organized as follows: 
In Section~\ref{more_details_coma}, we provide the detailed derivation for Eq.~\eqref{eq:update}, and the detailed algorithm of CoMA. Section~\ref{detailsbaselines}, we present a comparison to EWC, and main weight-averaging baselines. Section~\ref{more_exps}, we report extra experiments about \method, the impact of the hyperparameter $\lambda$ to influence stability-plasticity trade-off, and we detail the baselines, and datasets used in our experiments.

\section{More details CoMA/CoFiMA}
\label{more_details_coma}
\subsection{Detailed Derivations}
\label{derivative}
In the main paper, in Section~\ref{sec:coma}, we explain that using $\lambda=\frac{1}{t}$ in Eq.~\eqref{eq:update} is equivalent to the arithmetic average with uniform weights as in Eq.~\eqref{eq:avg}. We substantiate this claim through an inductive proof, detailed in the ensuing paragraphs.

Let us denote by \(\bar{\theta}_t\) the arithmetic mean of \(t\) distinct model parameters \(\thetavect_1, \thetavect_2, \ldots, \thetavect_t\). Commencing with the base case where \(t = 1\), the equivalence is intuitively apparent as \(\bar{\thetavect}_1 = \thetavect_1\).

To validate the inductive step, we proceed with the subsequent mathematical derivations:
\begin{align}
\bar{\thetavect}_t &= \frac{1}{t} \sum_{k=1}^{t} \thetavect_k\notag\\
&= \frac{1}{t} (\thetavect_t +\sum_{k=1}^{t-1} \thetavect_k\notag) \\
&= \frac{1}{t} \thetavect_t + \frac{t-1}{t} \frac{1}{t-1} \sum_{k=1}^{t-1} \thetavect_k\notag\\
&= \frac{1}{t} \thetavect_t + \frac{t-1}{t} \bar{\thetavect}_{t-1}\notag\\
&= \lambda \thetavect_t + (1-\lambda) \bar{\thetavect}_{t-1}, \text{with~} \lambda= \frac{1}{t}.
\end{align}

\subsection{Detailed Algorithm}
\label{detailedalgo}
In this section, we provide the detailed implementation of \method algorithm (see Alg.~\ref{algorithm:cfmacomplete}). We utilize the same notations as in SLCA~\cite{SLCA} to describe \method's training process.

\method begins with inputs that include a pre-training dataset $D_{pt}$, training datasets $D_t$ for tasks $t = 1, ..., T$, and a network $M_{\theta}(\cdot)$ with parameters $~{\theta = \{\theta_{rps}, \theta_{cls}\}}$. The parameters $\theta_{rps}$ are initialized by pre-training on $D_{pt}$, and $\theta_{cls}$ are randomly initialized. For each task, the network is trained using cross-entropy loss on $D_t$, applying different learning rates $\alpha$ and $\beta$ to $\theta_{rps}$ and $\theta_{cls}$, respectively, until convergence. 

After training, Fisher Diagonal information $F_t$ is calculated for the current parameters $\theta_t$, followed by an update to new weights $\theta^{*}_t$. The Fisher information is recalculated for these updated weights as $F_t^*$. This process includes collecting and saving mean $\mu_c$ and covariance $\Sigma_c$ for features $F_c$ for each class $c$ in $C_t$. 

In the final stage, the algorithm performs classifier alignment~\cite{SLCA}, where features $\hat{F}_c$ are sampled from a normal distribution defined by $\mu_c$ and $\Sigma_c$. The classifier $h_{\theta_{cls}}$ is then trained with normalized logits computed from these features until convergence, completing the task-specific training cycle.

\begin{algorithm*}[ht]
\caption{\small \method}
\label{algorithm:cfmacomplete}
\begin{flushleft}
\textbf{Input:} Training dataset $\mathcal{D}_t$ for task $t=1,...,T$;  network $M_{\varphi}(\cdot)$ with pre-trained parameters $\thetavect_0^*$;

\end{flushleft}
\begin{algorithmic}[1]
\For{each task $t \in \{1,...,T\}$} 
 \State \textit{\#\#Train network $M_{\phi}$ on $\mathcal{D}_t$ until convergence:}
 \While{not converged} 
  \State Update $M_{\phi}$ using cross-entropy loss.
 \EndWhile
  
  \State \textit{\#\#Feature collection:}
  \For{each class $c \in C_t$}
    \State Collect feature set $F_c=[r_{c,1}, ... , r_{c,N_c}]$.
    \State Compute mean $\mu_c$ and covariance $\Sigma_c$ of $F_c$.
  \EndFor
      \State \textit{\#\#Weight update process:}
  \State Compute Fisher Information $\Fmat_t$ for $\thetavect_t$.
  \State Update weights $\thetavect^{*}_t$ as per Eq.7.
  \State Calculate Fisher Information $\Fmat_t^*$ for $\thetavect_t^*$.
  
\EndFor
  \State \textit{\#\#Classifier alignment:}
  \For{each class $c \in C_{1:T}$}
    \State Sample $\hat{F}_c$ from $\mathcal{N}(\mu_c, \Sigma_c)$.
  \EndFor
  \State \textit{\#\#Train classifier $h_{\theta_{cls}}$:}
  \While{not converged} 
   \State Compute and normalize logits $H_{1:T}$.
   \State Update $h_{\theta_{cls}}$ using normalized logits.
  \EndWhile
\end{algorithmic}
\end{algorithm*}

\section{Comparison with alternative methods}
\label{detailsbaselines}

\subsection{Comparison to EWC (Elastic Weight Consolidation)}
\label{vs_ewc}
\noindent \textbf{Elastic Weight Consolidation (EWC)~\cite{kirkpatrick2017overcoming}. }EWC leverages the Fisher information matrix as a regularization mechanism within the loss function, penalizing deviations from previously learned tasks' parameters. The loss function in EWC is defined as:
\begin{equation}
\mathcal{L}(\thetavect) = \mathcal{L}_B(\thetavect) + \sum_i \frac{\lambda}{2} F_i (\thetavect_i - \thetavect^*_{A,i})^2
\end{equation}
where $\mathcal{L}_B(\thetavect)$ is the loss for the current task B, $\lambda$ represents the trade-off between learning new tasks and retaining old knowledge, $F_i$ denotes the Fisher information matrix elements for the previous tasks (that we name A), and $\thetavect^*_{A,i}$ are the optimal parameters for A.\\~\\

\noindent \textbf{Comparison and Discussion.}
\begin{itemize}
    \item \textbf{Utilization of Fisher Information. }EWC employs the Fisher information to inform the penalty within the loss function, thereby retaining critical features from previous tasks. Conversely, CoFiMA uses the Fisher information to modulate the parameter averaging, facilitating a balanced integration of knowledge from consecutive tasks.

    \item \textbf{Objective Function Implications. }The loss function in EWC is augmented with a regularization term, making the optimization process more complex. In contrast, CoFiMA's weighted averaging does not alter the loss function, potentially leading to a simpler optimization landscape.

    \item \textbf{Memory and Computational Requirements. }EWC necessitates storage of the entire history of Fisher information matrices and parameters, scaling with the number of tasks. CoFiMA, on the other hand, might only require the Fisher information matrix for the current and immediately preceding task, offering a more memory-efficient alternative.
    
\end{itemize}
\noindent 

Both EWC and CoFiMA offer innovative solutions to catastrophic forgetting. EWC emphasizes the preservation of performance on past tasks through penalization of parameter shifts, while CoFiMA promotes a dynamic balance across tasks via weighted averaging. The selection between these two approaches depends on the specific requirements of the application and the computational resources at hand.

\subsection{Comparison to Weight-Averaging Methods}
We provide an in-depth analysis of existing weight-averaging methods, including a comparative discussion.
\begin{itemize}
    \item \textbf{Weight-Ensemble}~\cite{wortsman2022model}: The key idea of this work is that by averaging the weights of multiple models that have been fine-tuned with different hyperparameter configurations, it is possible to improve both the accuracy and robustness of the resulting model. This method is referred to as "model soups". One of the main benefits is that, unlike conventional ensemble methods, averaging many models in this way does not incur additional inference or memory costs.

    \method differs from the Model Soup method by integrating Fisher information into the model averaging process, assigning weights to parameters based on their estimated importance to mitigate catastrophic forgetting in continual learning tasks. In contrast, Model Soup~\cite{wortsman2022model} averages weights of multiple fine-tuned models to improve accuracy without regard to the sequential nature of data or the forgetting of previous tasks.
    
    \item \textbf{WiSE-FT}~\cite{wortsman2021robust} introduces a method for enhancing robustness by ensembling the weights of the zero-shot and fine-tuned models. This approach aims to balance the need for high accuracy on target distributions while preserving the model's ability to perform well on varied and unforeseen data distributions as follows: $\thetavect_{\text{WiSE-FT}} = \alpha \thetavect_{\text{zero-shot}} + (1 - \alpha) \thetavect_{\text{fine-tuned}}$.

    \method utilizes Fisher information to dynamically weight parameter averaging across tasks, thereby mitigating catastrophic forgetting. In contrast, WiSE-FT focuses on robust fine-tuning of zero-shot models by interpolating between zero-shot and fine-tuned model weights to enhance robustness against distribution shifts without additional computational overhead.
    
    \item \textbf{Exponential Moving Average (EMA)}~\cite{szegedy2015rethinking}: EMA in deep learning models is a technique used to smooth out data by giving more weight to recent observations. It is often used in training neural networks to stabilize and improve the learning process by averaging the parameters over time: $\thetavect_{m} = \beta \thetavect_{m} + (1 - \beta) \thetavect_{m-1}$. In our setting, we found that the best performance is achieved by using $\beta=0.999$ on the evaluated datasets.

    Simon~\etal~\cite{simon2022generalizing} have already used EMA in the setting of continual learning in the context of knowledge distillation. Other works such as~\cite{pham2021continual, Zhang_2020_WACV} used EMA to stabilize training and reduce variance.

    \method differentiates itself from works~\cite{simon2022generalizing, pham2021continual, Zhang_2020_WACV} using EMA by leveraging Fisher information to adjust the averaging process, prioritizing parameters based on their relevance to preserve knowledge from previous tasks in CL. Furthermore, \method's averaging process is done after learning completely the task, not after some epochs. In contrast, EMA applies a decay factor to average model weights over time, primarily to stabilize training and improve convergence, without a task-specific focus on mitigating forgetting.

\end{itemize}

\section{More Experimental details}
\label{more_exps}
We outline the implementation details, additional experimental results, and description of the methods against which our approach is evaluated, and the datasets used in evaluation:

\subsection{Implementation Specifics}
A pre-trained ViT-B/16 backbone is utilized for our method as in \cite{SLCA}. For approaches that maintain fixed backbones and employ prompting, an Adam optimizer is used~\cite{wang2022dualprompt, wang2022learning} as in the original papers; conversely, for baselines that involve updating the entire model, an SGD optimizer is employed, both using a uniform batch size of 128. Our approach employs a learning rate of 0.0001 for the PTM and 0.01 for the classification layer. Empirically, Zhang~\etal~\cite{SLCA} found that supervised pre-training tends to reach convergence more rapidly than self-supervised pre-training. Thus, for supervised pre-training, we train all baselines for 20 epochs on CIFAR-100 and 50 epochs on other datasets. For self-supervised pre-training, a training duration of 90 epochs is applied across all benchmarks.

\subsection{Analysing Catastrophic Forgetting via Weight Distance}
To further evaluate the effectiveness of CoFiMA in continual learning, as discussed in Sec.~\ref{method}, we examine how CoFiMA achieves a balance between plasticity and stability from a perspective of Weight Distance.

\textbf{Weight Distance. }In addressing the issue of catastrophic forgetting, we prioritize the stability of model parameters as a key indicator of memory retention. This is grounded in the insight that lesser deviation from a model's previous weight configuration, upon incorporating new data, signals reduced forgetting~\cite{titsias2020functional, wang2023comprehensive}. 
In prior studies~\cite{ren2024analyzing}, the weight distance metric is utilized to quantify forgetting. This metric is defined as the $L_2$ norm of the difference between the model parameters after learning two consecutive tasks, expressed as:
\begin{equation}
\mathcal{W}(\thetavect_t,\thetavect_{t-1}) = \|\thetavect_t - \thetavect_{t-1}\|_2,
\end{equation}
where $\thetavect_t$ and $\thetavect_{t-1}$ represent the model parameters at the completion of tasks $t$ and $t-1$, respectively.

A strong advantage of CoMa is that we can demonstrate that, when learning a task $t$, starting from $\thetavect^*_{t-1}$, the update rule of CoMa guarantees a reduced weight distance compared to directly using the parameters $\thetavect_t$ learned on the task $t$:
\begin{align}
    \mathcal{W}(\thetavect^*_t, \thetavect^{*}_{t-1}) &= ||\thetavect^*_t - \thetavect^{*}_{t-1}||_2\notag\\
                                    &= ||\lambda\thetavect_{t}+(1-\lambda)\thetavect^*_{t-1}-\thetavect^{*}_{t-1}||_2\notag\\
    &= ||\lambda(\thetavect_{t}-\thetavect^*_{t-1})||_2\notag\\
    &= \lambda \mathcal{W}(\thetavect_t, \thetavect^{*}_{t-1})\notag\\
    &\leq\mathcal{W}(\thetavect_t, \thetavect^{*}_{t-1})~\text{since}~\lambda \in[0,1].
\end{align}

\begin{figure}[t]
\centering
\begin{subfigure}[t]{0.48\textwidth}
    \centering
    \includegraphics[width=\textwidth]{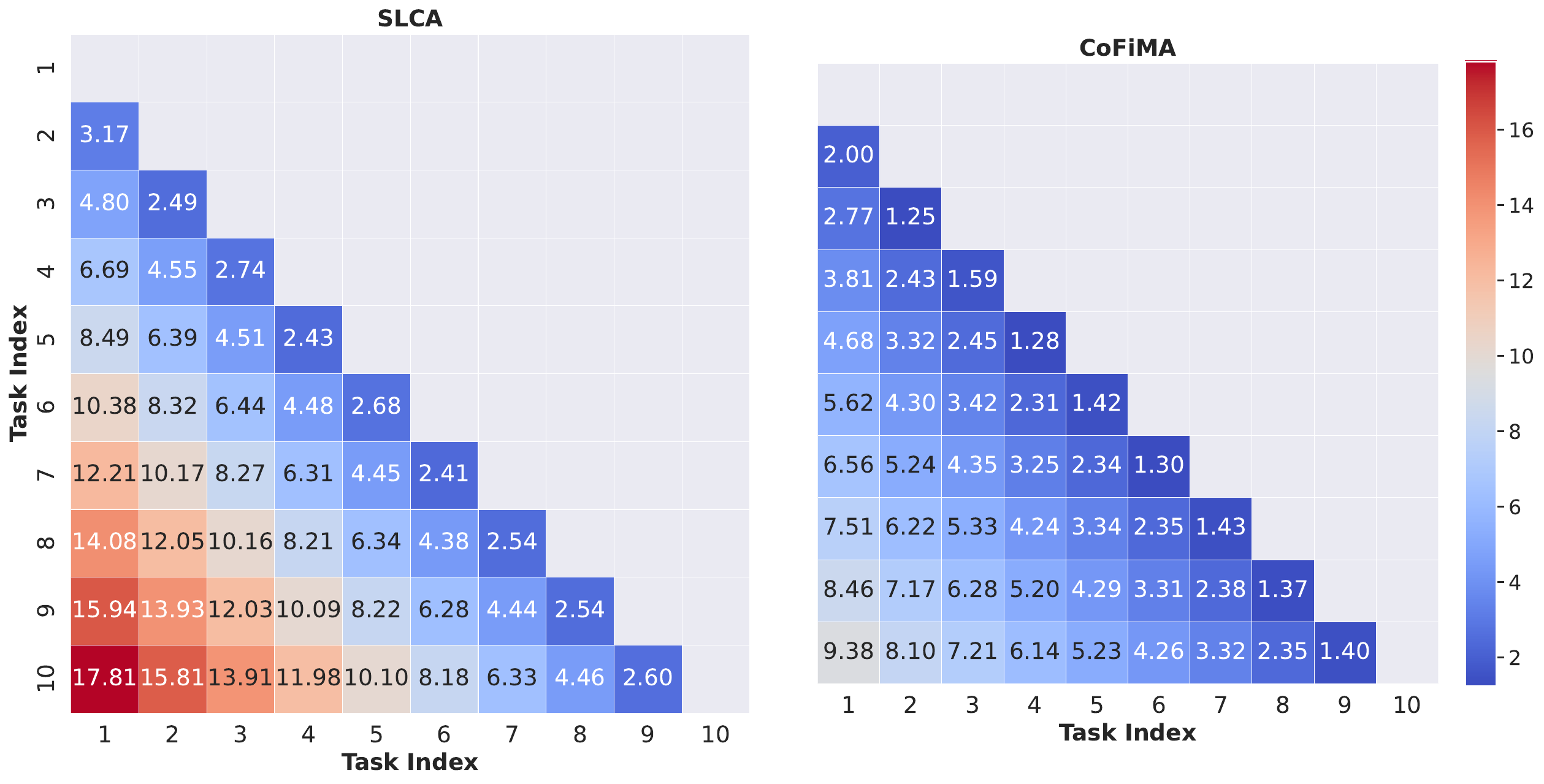}
    \caption{CIFAR-100}
    \label{fig:cifar_cofima}
\end{subfigure}
\begin{subfigure}[t]{0.48\textwidth}
    \centering
    \includegraphics[width=\textwidth]{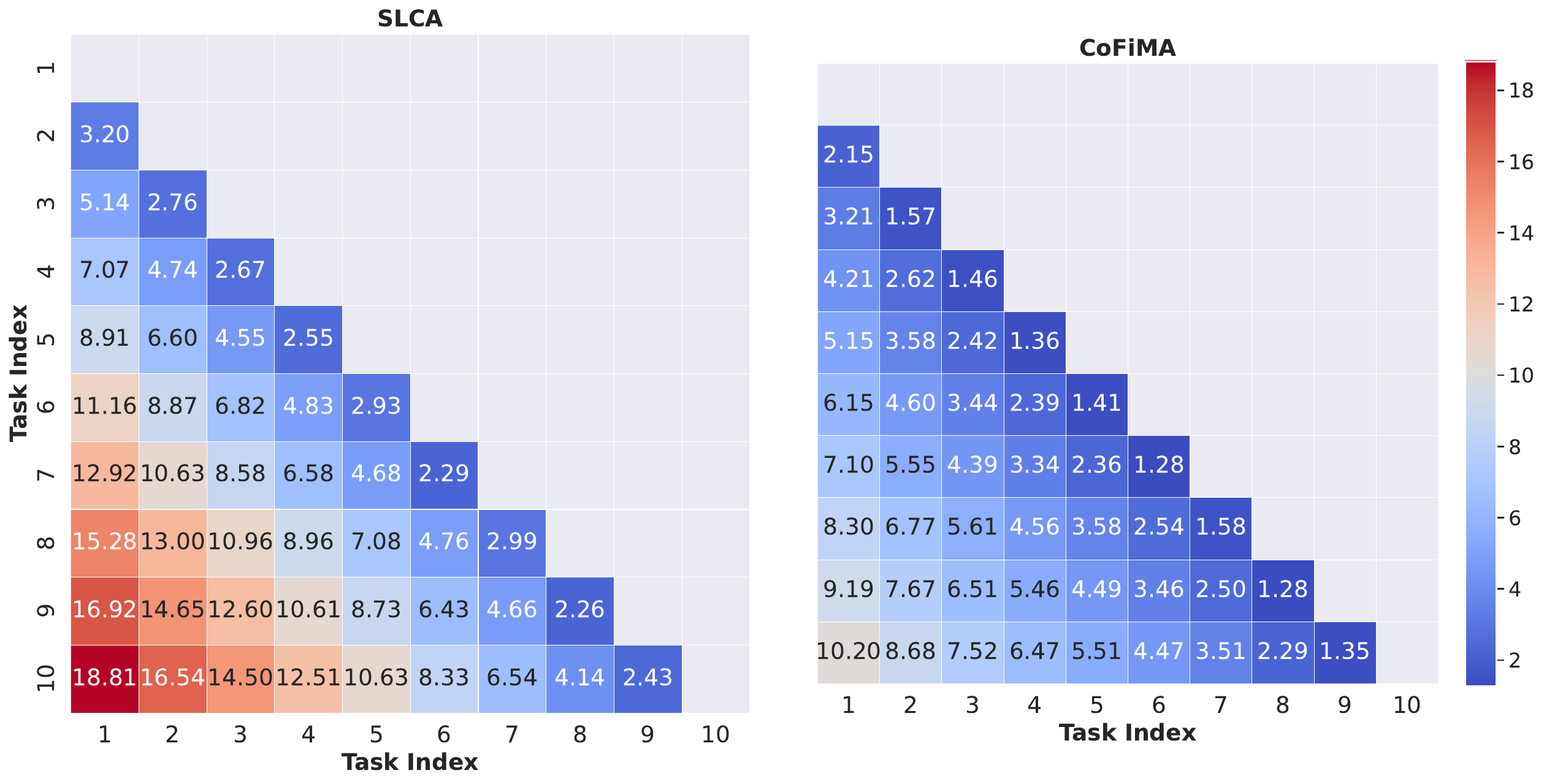}
    \caption{Imagenet-R}
    \label{fig:cub_slca}
\end{subfigure}
\caption{Similarity scores using $L_2$ norm for CIFAR-100 (a) and Imagenet-R (b) datasets using CoFiMA and SLCA methods.}
\label{fig:all_checkpoints}
\end{figure}
Note that similar derivations also hold for \method. To numerically assess the impact of this reduced weight distance, we report the weight distance $\mathcal{W}$ measured after every task in Fig.~\ref{fig:all_checkpoints} on two benchmarks. CoFiMA consistently exhibits lower weight distances across all checkpoints when compared to SLCA, indicating a more stable parameter retention throughout the learning process. Specifically, CoFiMA maintains a remarkably low Weight Distance, with values such as 2.00 and 1.67 for CIFAR-100 and CUB-200 respectively, compared to SLCA's higher values of 3.17 and 2.86 for the same datasets, suggesting a greater degree of parameter shift and, consequently, a higher susceptibility to forgetting. 

\subsection{Ablation on the Effect of $\lambda$}
\label{explit_coma}
As detailed in the main paper, the CoMA update rule is given by:
\begin{equation}
\thetavect^{*}_{t} = \lambda \thetavect_{t} + (1 - \lambda)\thetavect^*_{t-1},\label{eq:updatesupmat}
\end{equation}
where \( \thetavect^{*}_{t} \) is the parameter after CoMA at task \( t \), \( \thetavect_{t} \) is the parameter after training on task \( t \), and \( \lambda \) is the hyper-parameter that controls the blending of current and past parameters.
By considering $\thetavect^{*}_{1} = \thetavect_{1}$  for task \( t = 1 \), 
and iteratively applying Eq \eqref{eq:updatesupmat}, we obtain: 
\begin{equation}
\thetavect^{*}_{t} = \lambda \thetavect_{t} + \lambda(1 - \lambda) \thetavect_{t-1} + \lambda(1 - \lambda)^2 \thetavect_{t-2} + \ldots + \lambda(1 - \lambda)^{t-2} \thetavect_{2} + (1 - \lambda)^{t-1} \thetavect_{1},
\end{equation}

\begin{figure}[t]
    \centering
    \begin{minipage}[b]{0.48\textwidth} 
        \centering
        \includegraphics[width=\linewidth]{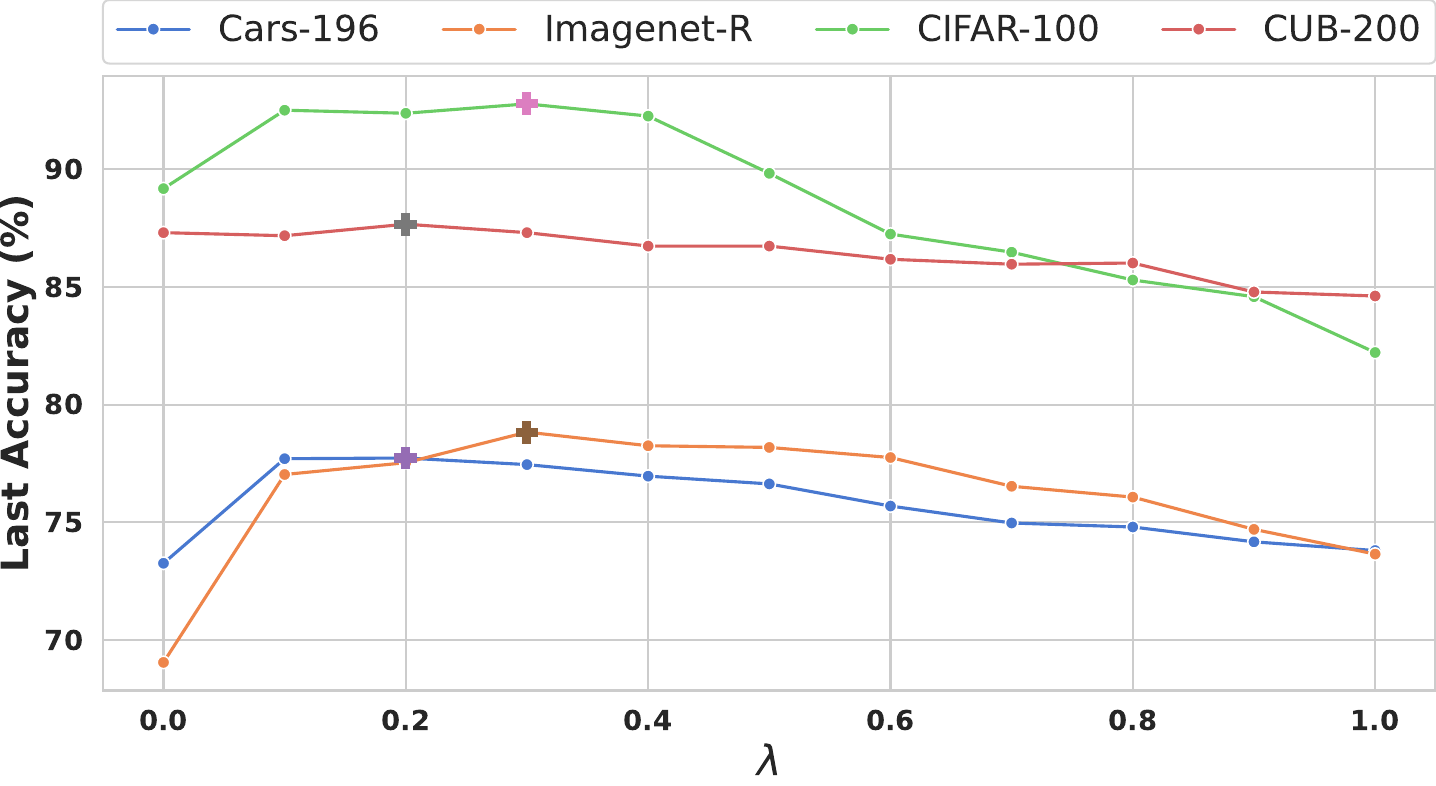}
        \label{fig:lambda_vit_side}
    \end{minipage}%
    \begin{minipage}[b]{0.48\textwidth} 
        \centering
        \includegraphics[width=\linewidth]{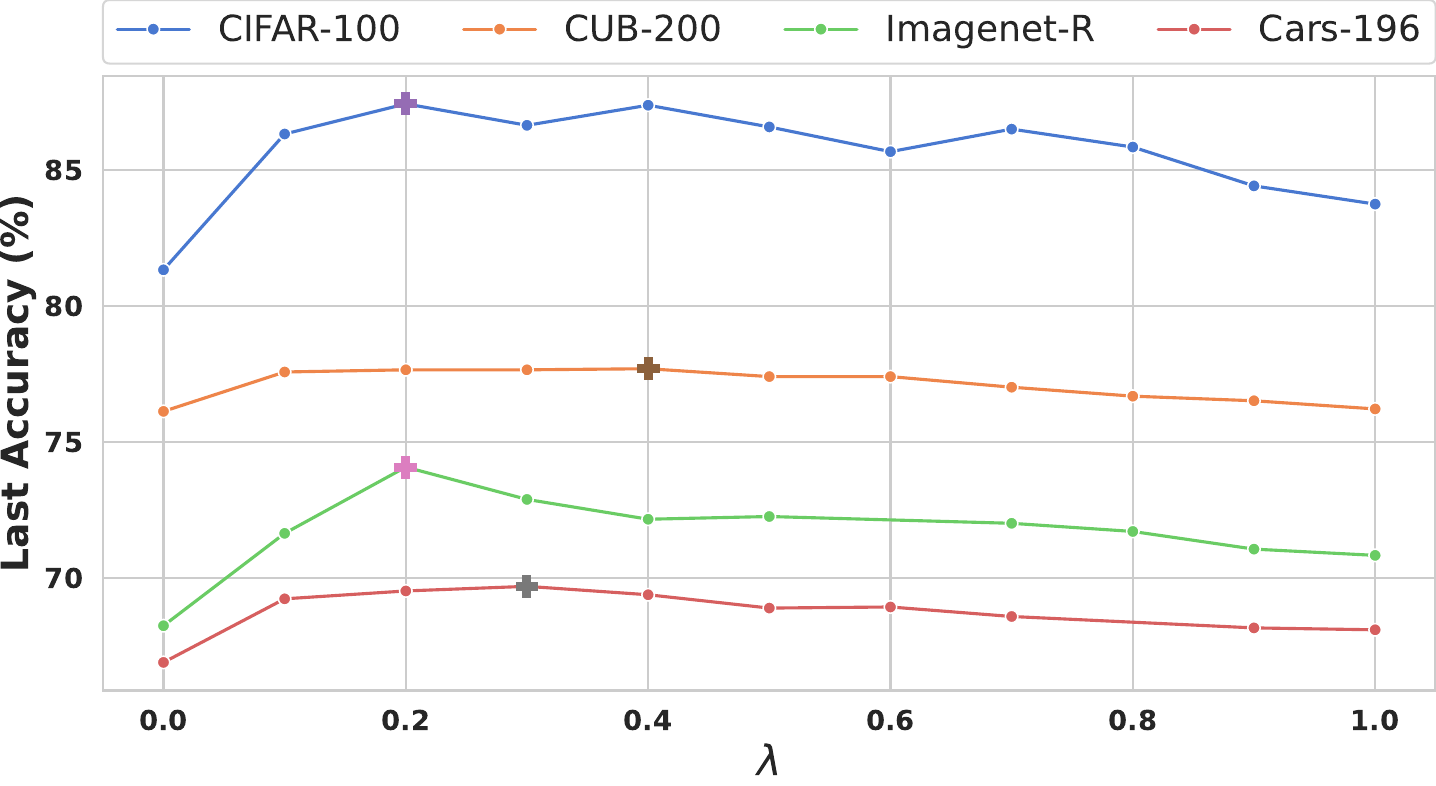}
        \label{fig:lambda_moco_side}
    \end{minipage}
    \caption{Assessing performance of \method using ViT-B/16 (\textbf{\textit{Left}}) and MoCo-V3 (\textbf{\textit{Right}}) with different values for $\lambda$. The marker $+$ represents the best performance.}
    \label{fig:lambda_vit_moco}
\end{figure}
\noindent which can be written as follows:
\begin{equation}
\thetavect^{*}_{t} = \lambda \sum_{k=1}^{t} (1 - \lambda)^{t-k} \thetavect_{k} + (1 - \lambda)^{t-1} \thetavect_{1}.
\label{eq_final_t}
\end{equation}

The update process for CoMA over multiple tasks unfolds in a recursive fashion, similarly to EMA, but with the distinction that the averaging is done over tasks (or classes in class-incremental learning) rather than iterations within a task. Eq.~\eqref{eq_final_t} shows that each past task's parameters are weighted by an exponentially decreasing factor (for a positive $\lambda < 1$) and the distance (in terms of tasks) from the current task $t$. This ensures that more recent tasks have a greater influence on the current parameter set than older tasks, which aligns with the goals of class-incremental learning where the model should adapt to new classes while still remembering the old ones.

Fig.~\ref{fig:lambda_vit_moco} presents the ablation study results for the CoFiMA method, focusing on the impact of the hyperparameter $\lambda$ on model performance across CIFAR-100, CUB-200, ImageNet-R, and Cars-196 datasets. The parameter $\lambda$ regulates the balance between stability and plasticity—retaining previous knowledge versus adapting to new tasks. For ViT-B/16, performance on CIFAR-100 and CUB-200 improves with increasing $\lambda$ to a point, then levels off or drops slightly, indicating an optimal $\lambda$ range for balancing new and old learnings. However, ImageNet-R and Cars-196 show a decline in performance beyond a certain $\lambda$ value, suggesting a need for greater focus on preserving past knowledge. Conversely, with MoCo-V3, CIFAR-100 benefits significantly as $\lambda$ approaches 1, highlighting the advantage of prioritizing recent learning. CUB-200 shows a positive trend up to a certain $\lambda$, after which performance drops, while Cars-196 remains stable across $\lambda$ values, indicating a different balance requirement. ImageNet-R performance worsens with higher $\lambda$, underscoring the importance of maintaining past knowledge for this dataset.

\subsection{Ablation Backbones}
\label{backbonesextra}
\begin{figure}[ht]
    \begin{center}
    \includegraphics[width=\columnwidth]{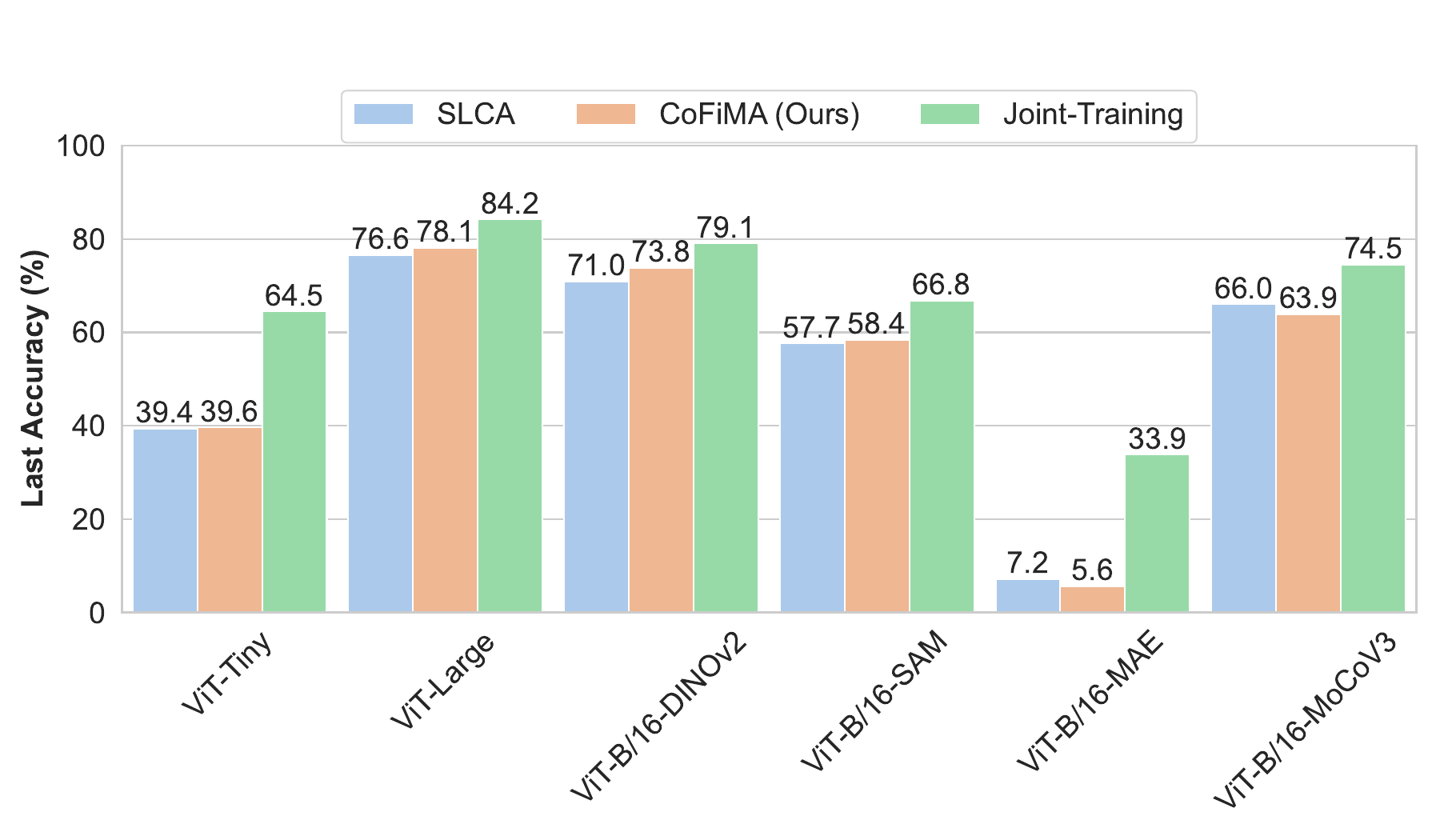}
    \end{center}
    \caption{Assessing performance of \method using various PTMs on CUB-200 dataset. \method enhances the results of SLCA.}
    \label{fig:ptms_cifar2}
\end{figure}

In this section, we evaluate the performance of \method approach across a variety of backbone architectures on another dataset CUB-196. Backbones evaluated are self-supervised (MAE~\cite{he2021masked}, MoCoV3~\cite{chen2021mocov3}, and DINOv2~\cite{dinoV2}) and supervised (ViT-Tiny~\cite{dosovitskiy2021image}, and ViT-B/16-SAM~\cite{foret2021sharpnessaware}) models. This analysis aims to evaluate the adaptability and performance consistency of \method with different architectures. Results are visualized in Fig.~\ref{fig:ptms_cifar2}.

Our results on CUB-200 dataset indicate that \method consistently enhances performance across all tested backbone architectures relative to the baseline SLCA method. Furthermore, the performance gain varies with the choice and the size of backbone, and its pertaining paradigm.

\subsection{Training from Scratch. }
\label{fromscratch}
We have conducted additional experiments to demonstrate the importance of starting from a PTM for the effectiveness of weight averaging. We evaluate the performance of Resnet18, and ViT-Tiny on CIFAR-100, and ImagenetR.

\begin{table}[ht]
\centering
\caption{Evaluation of \method using with Resnet18/ViT-T models trained from scratch.}
\begin{tabular}{c  ccccc}
\toprule
\multirow{1}{*}{} &  & \multicolumn{2}{c}{CIFAR100} & \multicolumn{2}{c}{ImageNet-R} \\ 
                       & & Scratch         & Pre-trained         & Scratch         & Pre-trained        \\ \midrule
\multirow{2}{*}{Resnet18}  & \multicolumn{1}{c}{Joint Training} & 69.05\tiny{$\pm 0.10$} & 76.12\tiny{$\pm 0.16$} &  63.64\tiny{$\pm 0.07$} & 71.49\tiny{$\pm 0.15$} \\ 
                           & \multicolumn{1}{c}{CoFiMA} & 37.24\tiny{$\pm 0.11$} & 68.31\tiny{$\pm 0.08$} & 48.92\tiny{$\pm 0.19$} &  68.89\tiny{$\pm 0.07$} \\ \midrule
\multirow{2}{*}{ViT-Tiny} & \multicolumn{1}{c}{Joint Training} & 63.15\tiny{$\pm 0.26$}  & 67.84\tiny{$\pm 0.14$} & 66.87\tiny{$\pm 0.38$} & 73.17\tiny{$\pm 0.19$} \\ 
                          & \multicolumn{1}{c}{CoFiMA} & 46.17\tiny{$\pm 0.17$} & 65.90\tiny{$\pm 0.23$} &  53.82\tiny{$\pm 0.04$} & 71.58\tiny{$\pm 0.27$} \\ \hline
\end{tabular}

\end{table}

We observe that CoFiMA's performance is low when we leverage models trained from scratch. We hypothesis that this due to starting from randomly initialized models leads to divergent solutions. Thus, averaging the weights might leads to high-loss region. This behavior has been investigated also by Neyshabur~\etal work ~\cite{neyshabur2021transferred}. They demonstrates that there exists a connection between minima obtained by pre-trained models versus freshly initialized ones. They observe that there is no performance barrier between solutions coming from pre-trained models, but there can be a barrier between solutions of different randomly initialized models, which suggests that the pre-trained weights guide the optimization to a flat basin of the loss landscape. On the other hand, barriers are clearly observed between the solutions from two instances trained from randomly initialized weights, even when the same random weights are used for initialization.

\subsection{Baselines}
\begin{itemize}
    \item \textbf{Finetune}: Incrementally trains on new datasets, inducing catastrophic forgetting as a consequence.
    \item \textbf{DER++~\cite{buzzega2020dark}}: mitigates catastrophic forgetting by using a replay buffer of past network inputs and predictions, combined with regularizing current predicted logits, effectively blending rehearsal, knowledge distillation, and regularization.
    \item \textbf{LwF~\cite{li2017learning}}: Employs knowledge distillation~\cite{hinton2015distilling} as a regularizer to mitigate forgetting, relying on the legacy model for soft target generation.
    \item \textbf{L2P~\cite{wang2022learning}}: A leading PTM-based CIL method that maintains a frozen pre-trained model while optimizing a prompt pool. It incorporates a 'key-value' pairing mechanism for prompt selection and leverages an auxiliary pre-trained model for prompt retrieval.
    \item \textbf{DualPrompt~\cite{wang2022dualprompt}}: An extension of L2P that utilizes two categories of prompts—general and expert—for enhanced performance. It also uses an additional pre-trained model for prompt retrieval.
    \item \textbf{SLCA~\cite{SLCA}}: improves the classification layer by modeling the class-wise distributions and aligning the classification layers in a post-hoc fashion.
    \item \textbf{RanPAC~\cite{mcdonnell2024ranpac}}: injects a frozen Random Projection layer with nonlinear activation between the PTM's feature representations and output head, which captures interactions between features with expanded dimensionality, providing enhanced linear separability for class-prototype-based CL.
    
\end{itemize}

\subsection{Datasets}
\label{datasets}

\begin{itemize}
    \item \textbf{CIFAR100}~\cite{krizhevsky2009learning}: Comprises 100 classes, 60,000 images 50,000 for training, and 10,000 for testing.
    \item \textbf{CUB200}~\cite{WahCUB2002011}: Focuses on fine-grained visual categorization, containing 11,788 bird images across 200 subcategories, with 9,430 for training and 2,358 for testing.
    \item \textbf{ImageNet-R}~\cite{hendrycks2021many}: Extended for CIL by~\cite{wang2022dualprompt}, includes various styles and hard instances, totaling 24,000 training and 6,000 testing instances.
    \item \textbf{Cars-196}~\cite{stanfordcars}: is a collection of over 16,000 images of 196 classes of cars, categorized by make, model, and year. It is commonly used for fine-grained image recognition and classification tasks in machine learning.
\end{itemize}

\begin{table}[t]
	\caption{Introduction about benchmark datasets.
		ObjectNet, OmniBenchmark, and VTAB contain massive classes, and we sample a subset from them to construct the incremental learning task.}
	\label{tab:supp_dataset}
	\centering
		\begin{tabular}{lcccccc}
			\toprule
			Dataset & \# training  & \# testing  & \# Classes& Link \\ 
            &instances&instances\\\midrule
			CIFAR100 & 50,000 & 10,000 & 100 & \href{https://www.cs.toronto.edu/~kriz/cifar.html}{\nolinkurl{Link}}\\
			CUB200 & 9,430 & 2,358 & 200 & \href{https://www.vision.caltech.edu/datasets/cub_200_2011/}{\nolinkurl{Link}}\\
			ImageNet-R & 24,000 & 6,000 & 200 & \href{https://github.com/hendrycks/imagenet-r}{\nolinkurl{Link}}\\
                Cars196 & 8,144 & 8,044 & 196 & \href{https://github.com/hendrycks/imagenet-r}{\nolinkurl{Link}}\\
     
		VTAB & 1,796 & 8,619 & 50 & \href{https://google-research.github.io/task_adaptation/}{\nolinkurl{Link}}\\
			\bottomrule
		\end{tabular}
\end{table}

\end{document}